\documentclass[10pt,twocolumn,letterpaper]{article}

\usepackage{cvpr}              % To produce the CAMERA-READY version
\usepackage[dvipsnames]{xcolor}
\usepackage{multirow}
\usepackage{tikz}
\usepackage{pgfplots}
\usepackage{colortbl}
\usepackage{graphicx} 
\usepackage{bm}
\usepackage{microtype}
\usepackage{amsmath}
\usepackage{appendix}
\usepackage{float}
\definecolor{cvprblue}{rgb}{0.21,0.49,0.74}
\usepackage[pagebackref,breaklinks,colorlinks,citecolor=cvprblue]{hyperref}
\usepackage{array}

\usepackage{algorithm}
\usepackage{algorithmicx}
\usepackage{algpseudocode}
\usepackage{amsmath}

\usepackage{booktabs}
\usepackage{bbding}
\usepackage{makecell}
\usepackage{microtype}

\usepackage{colortbl}
\definecolor{mygray}{gray}{.9}

\def\paperID{2571} % *** Enter the Paper ID here
\def\confName{CVPR}
\def\confYear{2024}

\title{Learning to Transform Dynamically for Better Adversarial Transferability}
\author{
\thanks{Equal contribution}~~\thanks{Project lead}~~Rongyi Zhu \quad $^{*\dag}$Zeliang Zhang \quad Susan Liang \quad Zhuo Liu \quad \thanks{Corresponding author}~~Chenliang Xu\\
University of Rochester\\
{\tt\small \{rongyi.zhu, zeliang.zhang, susan.liang,  zhuo.liu, chenliang.xu\}@rochester.edu}}
\begin{document}
\maketitle
\begin{abstract}
%Adversarial examples are human-imperceptible perturbations on inputs to fool the neural network, which also present the phenomena of the adversarial transferability among different models. Input transformation-based methods improve the attack performance by diversifying the inputs, but further improvements are limited by fixed single transformations and dynamic changes of optimal combinations of transformations. In this work, we identify that there exists an optimal combination of input transformation-based methods to boost the adversarial transferability in each iteration. We propose a novel approach  (L2T) to learn the dynamic optimal combination to augment the images by maximizing the input diversity for better attack performance. We evaluate our proposed L2T on the ImageNet-1K dataset, competing with various methods and against many defense methods. We also assess the performance of our method by the real-world application Google's Vision API. We achieve the state-of-art performance on all tasks.   

Adversarial examples, crafted by adding perturbations imperceptible to humans, can deceive neural networks. Recent studies identify the adversarial transferability across various models, \textit{i.e.}, the cross-model attack ability of adversarial samples. To enhance such adversarial transferability, existing input transformation-based methods diversify input data with transformation augmentation. However, their effectiveness is limited by the finite number of available transformations. In our study, we introduce a novel approach named Learning to Transform (L2T). L2T increases the diversity of transformed images by selecting the optimal combination of operations from a pool of candidates, consequently improving adversarial transferability. We conceptualize the selection of optimal transformation combinations as a trajectory optimization problem and employ a reinforcement learning strategy to effectively solve the problem. Comprehensive experiments on the ImageNet dataset, as well as practical tests with Google Vision and GPT-4V, reveal that L2T surpasses current methodologies in enhancing adversarial transferability, thereby confirming its effectiveness and practical significance. The code is available at \url{https://github.com/RongyiZhu/L2T}.

\end{abstract}    
\section{Introduction}
\label{sec:intro}

Neural networks have been adopted as the building block for various real-world applications, such as face detection~\citep{florian2015facenet,wang2018cosface, liu2024emo}, autonomous driving~\citep{geiger2012we,lillicrap2015continuous}, and medical diagnosis~\citep{bakator2018deep,richens2020improving}. However, neural networks are vulnerable to adversarial examples, which contain human imperceptible adversarial perturbations on the benign input. This issue is increasingly concerning researchers, as it is essential for ensuring the trustworthy use of neural networks~\citep{chatila2021trustworthy, zhang2024random,jiang2023one,zhang2023novel, zhang2023towards, zhang2024discover, zhang2024forward}.

%among others, the issue of AI security is increasingly concerning researchers, as it is essential for ensuring their trustworthy use~\citep{chatila2021trustworthy}. The adversarial attack~\citep{szegedy2014intriguing} is one of the most powerful techniques to threaten the reliable use of AI. Adversarial attack crafts human imperceptible adversarial perturbations on the benign input to fool the deep neural networks (DNNs). The study of advanced adversarial attacks helps understand the DNNs~\citep{ma2021understanding}, as well as improves the robustness of DNNs~\citep{yan2018deep,guo2018sparse}. 

\begin{figure}
    \includegraphics[width=\linewidth]{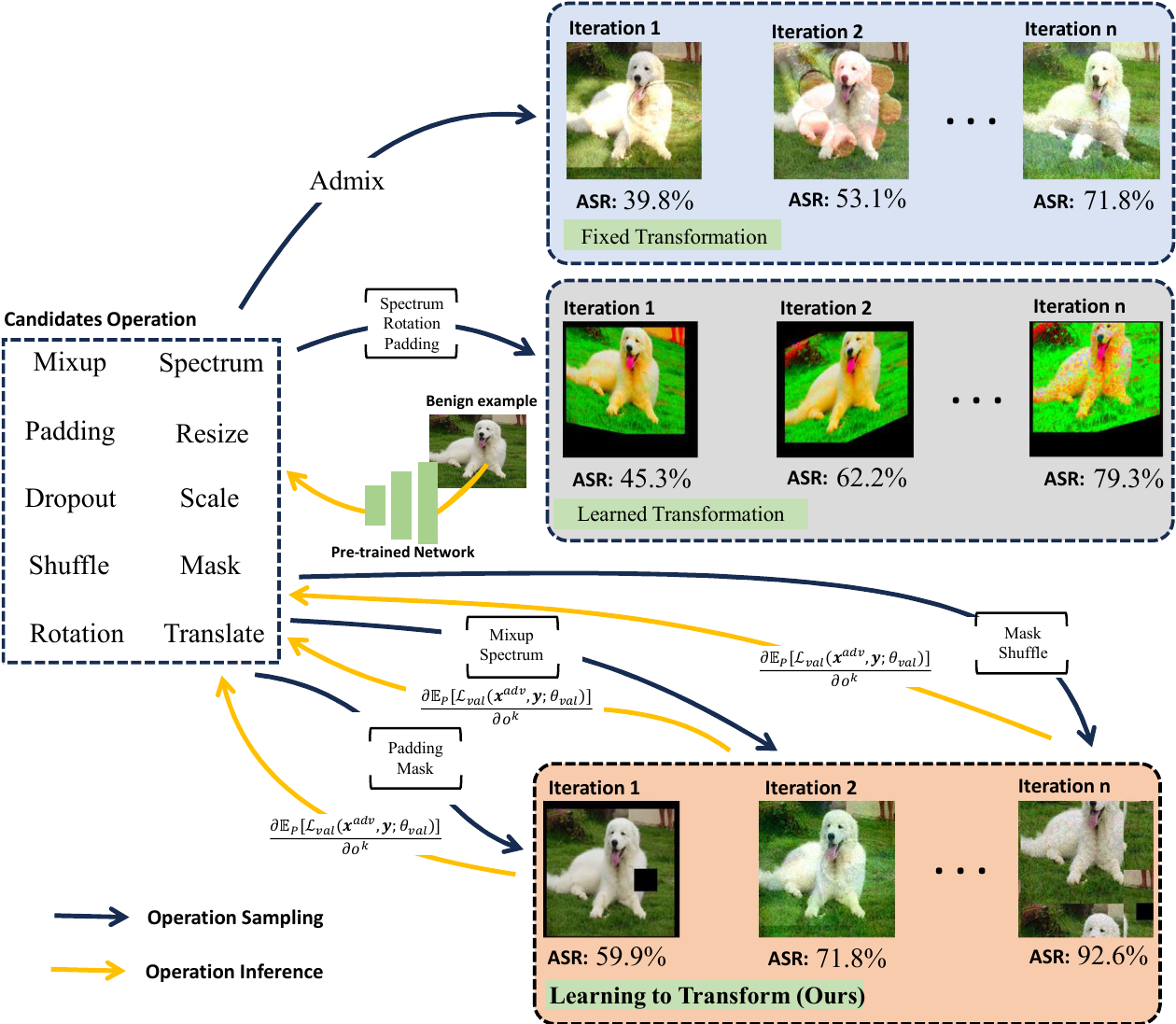}
    \caption{For input transformation-based attacks, most works design a fixed transformation and use it to craft the adversarial perturbation. The learning-based methods preliminarily predict augmentation strategies for current images for better adversarial transferability. These methods cannot respond to the distribution shifts between benign images and adversarial examples. We propose Learning to Transform (L2T), which uses the dynamic of the optimal transformation in each iteration to further boost the adversarial transferability. }
    \label{fig:teaser}
\end{figure}

In real-world scenarios of adversarial attacks~\citep{wu2021adversarial, tang2004video, liu2023differentially}, the target model is usually inaccessible. To attack these inaccessible models, many studies instead rely on surrogate models to generate adversarial examples~\citep{xie2019improving,dong2018boosting,zhang2024bag} and use generated samples to mislead the target model. This cross-model attack ability of samples generated on the surrogate models is called ``adversarial transferability.'' Numerous research studies are dedicated to enhancing adversarial transferability, which can be classified into four categories: gradient-based methods~\citep{dong2018boosting,lin2019nesterov,wang2021enhancing,wang2021boosting}, input transformation-based methods~\citep{xie2019improving,dong2019evading,lin2019nesterov,wang2021admix}, architecture-based methods~\citep{li2020learning,wu2020skip}, and ensemble-based methods~\citep{liu2016delving, xiong2022stochastic}. Among these attack methodologies, input transformation-based methods gain much popularity because of their plug-n-play advantage, which can be seamlessly integrated into other attack techniques~\cite {wang2021enhancing, dong2018boosting}.  However, we discover that existing input transformation-based methods adopt the same transformation when crafting adversarial examples, limiting the flexibility of transformation operations. We hypothesize that we should select the optimal transformation dynamically in each iteration to enhance the adversarial transferability.  

As shown in Fig.~\ref{fig:teaser}, prior input transformation-based methods often revolve around designing fixed augmentation strategies like resizing inputs~\citep{xie2019improving}, block masking~\citep{fan2022maskblock}, or mix-up~\citep{wang2021admix}. A more dynamic approach is presented by~\citep{yuan2021automa}, advocating the precomputation of various sequences of augmentation strategies to apply to each iteration to enhance the attack performance. Complementing this, \citet{wu2021improving} proposes the use of generative models for image augmentation to boost the adversarial transferability. Some studies go further, combining multiple augmentation strategies to amplify input diversity to improve the performance. For example, \citet{yuan2022adaptive} introduces a neural network that generates a prediction of the optimal transformation strategy and applies the strategy to improve performance. A further improvement is hindered by the limited number of transformations.

% One way to improve the transferability is to apply transformations on the Image before forwarding to the network. Many works have studied how these transformations can influence the transferability of adversarial examples. SIM, Admix, and SSA found certain transformation strategies that enhance adversarial transferability. AutoMA, ATTA, and AITL utilize adaptive transformations to images. However, these methods still cannot answer the difference between iterations. We find that adjusting the combination of transformations in iterations has an impact on the transferability. Based on this fact, we propose that there exists an optimal combination of input transformations for different iterations in the generation of adversarial examples.

To fully utilize the limited number of transformations, a natural idea is to use a combination of operations. However, it is not always efficient to combine different transformations together for attack, as reported in \citep{wang2023structure}. We expect to find an optimal combination of transformations to achieve a trade-off between operation diversity and adversarial transferability. Nonetheless, the enormity of the search space presents a significant challenge, impeding the identification of the most efficacious combination of transformations during an attack for optimal adversarial transferability. To surmount this hurdle, we conceptualize the search process of the optimal combination of transformations as a problem of optimal trajectory search. Each node within this trajectory represents an individual transformation, and each directed edge means a transfer of the optimal transformation from the current step to the next step. To effectively obtain the optimal trajectory in such a large search space, we design a reinforcement learning-based approach, capitalizing on its demonstrated efficacy in navigating expansive search domains.

% We believe that establishing a dynamic optimal transformation strategy is pivotal for enhancing adversarial transferability. There are two major challenges in this problem. \underline{First}, due to the enormous possible transformation, finding an optimal transformation for different image in each iteration is a time-consuming task. This makes it impossible to search for the optimal transformation for every image.  \underline{Second}, it is hard to define the dynamic goodness of different input transformation methods and apply the feedback to adjust strategies in time. \underline{Third}, prior knowledge of benign examples is not accessible in real-world practices. It is impossible to infer a solution from past experience. 

%it remains a question about how to better utilize the learned strategy to augment the images with maximum diversity for better performance.

In this paper, we introduce a novel framework called Learning to Transform (L2T) to improve the adversarial transferability of generated adversarial examples. L2T dynamically learns and applies the optimal input transformation in each iteration. Instead of exhaustively enumerating all possible input transformation methods, we employ a reinforcement learning-based approach to reduce the search space and better utilize the transformations to improve the diversity. In each iteration of the adversarial attack, we sample a subset of transformations and apply them to the adversarial examples. Subsequently, we update the sampling probabilities by conducting gradient ascent to maximize the loss. Our method effectively learns the dynamics of optimal transformations in attacks, leading to a significant enhancement in adversarial transferability. Additionally, compared to other learn-based adversarial attack methods, our approach is more efficient for adversarial example generation, as it obviates the need for additional training modules.

We summarize our contributions as follows,
\begin{itemize}
    \item We formulate the problem of optimal transformation in adversarial attacks, which studies finding the optimal combination of transformations to increase the input diversities, thus improving the adversarial transferability. 
    \item We propose Learning to Transform (L2T) that exploits the optimal transformation in each iteration and dynamically adjusts transformations to boost adversarial transferability.
    \item Extensive experiments on the ImageNet dataset demonstrate that L2T outperforms other baselines. We also validate L2T's superiority in real-world scenarios, such as Google Vision and GPT-4V.
\end{itemize}

%we choose an optimal combination of transformations for each iterations. By optimizing the transformations in each iteration, the transferablity of adversarial examples are also optimized. 

%3. Basic IDEA
% adaptive 
%In this paper, we proposed an direct differential approaches to approximate the optimal solution in input transformation combination problem. Our method 
%We first formulate the combination optimization problem. Our target is to find an optimal input transformation combination for adversarial transferability. This problem can be seen as a dual optimization problem. First, we need to an optimal optimal input transformation combination. Second, we need to find an optimal adversarial examples to evaluate the combination. 

%We proposed an approximate solution to input transformation combination problem. We relax an optimization process by one-step gradient approximation. Thus, the combination policy and adversarial examples are updated together. 

%The proposed method has three favorable features. First, our method do not need to training. Second, we do not need to obtain the prior knowledge to data distribution. Third, our method can corresponding to the iterative dynamic in generating adversarial attacks. Thus, our method have a stronger adaptation power to the difference among images. 

%4. Summary of work Extensive experiments on 
\section{Related Work}
\label{sec:Related_works}

\subsection{Adversarial Attack}
Various adversarial attacks have been proposed, \eg, gradient-based attack~\cite{goodfellow2014explaining,kurakin2018adversarial, madry2018towards}, transfer-based attack~\cite{dong2018boosting, xie2019improving,wei2019transferable,long2022frequency}, score-based attack~\cite{ilyas2018black,li2019nattack,chen2017zoo}, decision-based attack~\cite{brendel2017decision, li2020qeba,wang2021triangle}, generation-based attack~\cite{xiao2018generating,wang2019gan}. Among these, transfer-based attacks do not require the information of the victim models, making it popular to attack the deep models in the real world and raise more research interests. To improve adversarial transferability, various momentum-based attacks have been proposed, such as MI-FGSM~\cite{dong2018boosting}, NI-FGSM~\cite{lin2019nesterov}, VMI-FGSM~\cite{wang2021enhancing}, EMI-FGSM~\cite{wang2021boosting}, \etc. Several input transformation methods are also proposed, such as DIM~\cite{xie2019improving}, TIM~\cite{dong2019evading}, SIM~\cite{lin2019nesterov}, Admix~\cite{wang2021admix}, SIA~\cite{wang2023structure}, STM~\cite{ge2023improving}, BSR~\cite{wang2023rethinking}, \etc, which augment images used for adversarial perturbation computation to boost transferability.  The input transformation-based methods can be integrated into the gradient-based attacks for better performance.   

Delving into the input transformation-based methods, most works are limited to designing a fixed transformation to augment the images, which limits the diversity of transformed images and the adversarial transferability. To address this issue, some researchers~\cite {wu2021improving, yuan2022adaptive, yuan2021automa} propose to augment the images with a set of multiple transformations predicted by a pre-trained network.  Automatic Model Augmentation (AutoMA) \cite{yuan2021automa} adopts a Proximal Policy Optimization (PPO) algorithm in search of a strong augmentation policy. Adversarial Transformation-enhanced Transfer Attack (ATTA) \cite{wu2021improving} proposes to employ an adversarial transformation network in modeling the most harmful distortions. Adaptive Image Transformation Learner (AITL)~\cite{yuan2022adaptive} incorporates different image transformations into a unified framework to learn adaptive transformations for each benign sample to boost adversarial transferability. By applying optimal multiple transformations, the adversarial attack performance is largely improved.

\subsection{Adversarial Defense}
Various defense approaches have been proposed to mitigate the threat of adversarial attacks,  such as adversarial training~\cite{madry2018towards,Tramr2018EnsembleAT,wang2021multi}, input preprocessing~\cite{xie2018mitigating,Naseer_2020_CVPR}, feature denoising~\cite{liao2018defense,xie2019feature,yang2022robust}, certified defense~\cite{raghunathan2018certified,gowal2019scalable,cohen2019certified}, \etc.  \citet{liao2018defense} train a denoising autoencoder, namely the High-level representation guided denoiser (HGD), to purify the adversarial perturbations. \citet{xie2018mitigating} propose to randomly resize the image and add padding to mitigate the adversarial effect, namely the Randomized resizing and padding (R\&P).  \citet{xu2018feature} propose the Bit depth reduction (Bit-Red) method, which reduces the number of bits for each pixel to squeeze the perturbation.  \citet{liu2019feature}  defend against adversarial attacks by applying a JPEG-based compression method to adversarial images.  Cohen~\etal~\cite{cohen2019certified} adopt randomized smoothing (RS) to train
a certifiably robust  classifier.  \citet{Naseer_2020_CVPR} propose a Neural Representation Purifier (NRP)  to eliminate perturbation.

\section{Learning to Transform}
\label{sec:method}
%\textbf{Notations}: Scalars, vectors, matrices are denoted by lowercase, boldface, lowercase, boldface uppercase and calligraphic letters, respectively, e.g. $x \in \mathbb{R}$, $ \mathbf{x} \in \mathbb{R}^n$, $\mathbf{X} \in \mathbb{R}^{n_1 \times n_2}$, and $\mathcal{X} \in \mathbb{R}^{n_1 \times n_2 \times n_3}$.

\subsection{Task definition}

%We start from adversarial attack with input transformation. Given a natural sample $\bm{x}$ and the corresponding label $y$, an adversarial attack transforms the natural sample $\bm{x}$ into a adversarial sample $\bm{x}^{adv}$. The adversarial sample $\bm{x}^{adv}$ target to mislead a target classifier $f_t$ from the true label $y$, \ie, $f(\bm{x}^{adv}) \neq y$. For black-box attack, the targeted classifier $f_t$ is unknown. The attacker takes a surrogate classifier $f_{\theta}$ to craft adversarial sample $\bm{x}^{adv}$ and relies on the adversarial transferability to fool the targeted classifier $f_t$. In practice, attacker adopts gradient ascent to iteratively update $x^{adv}$ to maximize the loss between $f_{\theta}(x^{adv})$ and $y$. Takes I-FGSM as an example, the adversarial example $x^{adv}_{t}$ at iteration $t$ can be expressed as:

The crafting of adversarial examples usually takes an iterative framework to update the adversarial perturbation. Given a benign sample $\bm{x}$ and the corresponding label $y$, a transferable attack takes a surrogate classifier $f_{\bm{\theta}}$ and iteratively updates the adversarial example $\bm{x}^{adv}$ to maximize the loss of classifying $f_{\theta}(\bm{x}^{adv})$ to $y$. Take I-FGSM~\citep{song2018physical} as an example. The adversarial example $\bm{x}^{adv}_{t}$ at the $t$-th iteration can be formulated as follows:
\begin{equation}
\label{eq:adv_upt}
    \bm{x}^{adv}_{t} = \bm{x}^{adv}_{t-1} + \alpha \cdot {\text{sign} } (\nabla_{\bm{x}^{adv}_{t-1}} J(f_{\bm{\theta}}(\bm{x}^{adv}_{t-1}, y) ) ),
\end{equation}

\begin{align*}
\label{eq:adv_upt}
    \bm{x}^{adv}_{t} &= \bm{x}^{adv}_{t-1} \\  &+ \alpha \cdot {\text{sign} } (\nabla_{\bm{x}^{adv}_{t-1}} J(f_{\bm{\theta}}(\bm{x}^{adv}_{t-1}, y) ) ),
\end{align*}

where we denote $\alpha$ as the step size, $J(\cdot, \cdot)$ as the classification loss function. As identified by previous studies, the adversarial example exhibits a characteristic of transferability, where the adversarial examples generated by the surrogate model can fool other neural networks.

%Input transformation-based methods are one of the most effective methods to boost adversarial transferability. With these methods,  the adversarial samples are firstly transformed by a set of image transformations and then proceeded to gradient calculation. Let  $\varphi$ denote a set of image transformations $o$, where $\varphi = \{ o^i | i \in \{1,2,...,k\} \}$. At the $t$-th iteration,  the adversarial example $\bm{x}^{adv}_t$ is transformed sequentially by $o^i$ as follows,

Input transformation-based methods are one of the most effective methods to boost adversarial transferability. With these methods,  the adversarial samples are firstly transformed by a set of image transformations and then proceeded to gradient calculation. Let  $\varphi$ denote a set of image transformations operation $o$, where $\varphi = \{ o^i | i \in \{1,2,...,k\} \}$. At the $t$-th iteration,  the adversarial example $\bm{x}^{adv}_t$ is transformed sequentially by $o^i$ as follows,

\begin{equation}
    \varphi(\bm{x}^{adv}_t) = o^{k} \oplus o^{k-1} \oplus \cdots \oplus o^1 (\bm{x}^{adv}_{t}),
\end{equation}
where $o^2 \oplus o^1 (\bm{x})$ denotes the operation $o^2(o^1(\bm{x}))$, $o^{1}, o^{2} \in \varphi $. We use the gradient of $\varphi(\bm{x}^{adv}_t)$ with respect to the loss function to update the adversarial perturbation as  \cref{eq:adv_upt}.

There are two categories for selecting the operation set $\varphi$ in the previous study.   One line of research focuses on designing fixed transformation-based methods, which use a pre-defined transformation $\varphi$.   For example, Admix chooses mixup and scaling for transformation $\varphi$.   The other line of research proposes the learning-based transformation methods, which usually use a generative model to directly generate the transformed $\varphi(x)$. Compared with the fixed transformation-based methods, learning-based methods enjoy more diversity of transformed images, leading to a better performance in adversarial transferability. In our work, we study the learning-based transformation methods.

\subsection{Motivation}

%Since Kurakin \etc introduced to iteratively update the adversarial examples, many works have adopted iterations in their methods. As illustrated in Figure \ref{}, we examine the transferability of adversarial examples generated by different transformations.

\begin{figure}
    \centering
    \includegraphics[width=0.9\linewidth]{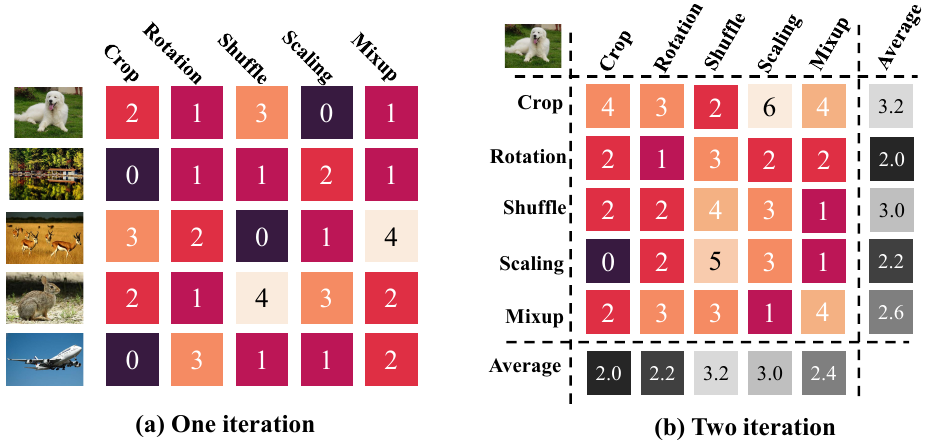}
    \caption{Comparsion for different operations in boosting the adversarial transferability. The number in the box denotes the number of fooled models (Maximum: 9). In (a), the horizontal axis denotes different transformation operations and the vertical axis denotes different benign examples. In (b), the vertical axis denotes the transformation used in the first iteration and the horizontal axis denotes the transformation used in the second iteration}
    \label{fig:iteraion1}
\end{figure}

% There are many works discussing what is a good transformation for adversarial transferability. To better understand the relationship between adversarial transferability and transformation. We proposed that we should investigate iteration to understand how transformation helps adversarial attacks to transfer. 

Previous research designs lots of transformations to improve the diversity of images, thus guiding the adversarial attacks to focus more on the invariant robust features. However, it does not always work by increasing the number of transformed images for attacks to boost the adversarial transferability. Because some combination of transformations can cause damage to original examples, losing massive amounts of information used for transferable attacks.  A natural question occurs to us, \textit{for one image, does there  exist the optimal combination of transformations for the best adversarial transferability?} 

To answer this question, we start by generating adversarial examples in one iteration. We take an example of crafting adversarial examples using ResNet-18 to attack other $9$ models\footnote{ResNet-101, DenseNet-121, ResNext-50, Inception-v3, Inception-v4, ViT, PiT, Visformer, Swin}. We denote $5$ operations for input transformation methods, namely the crop, rotation, shuffle, scaling, and mix-up. We use these operations on five images for attacks and report the number of models fooled. We report the results in \cref{fig:iteraion1}. It can be seen that by shuffle, we can achieve the maximum transferable attack success rates on a dog image, indicating the optimal transformations in all possible $5$ operations. 

We continue our discussion in the two-iteration scenario. Following the same setting in one iteration, we report the number of fooled models. It can be seen that by choosing crop in the first iteration and scaling for the second iteration, which successfully fooled $6$ models out of $9$. We also notice that shuffle, the optimal transformation in one iteration, can not maintain the optimal performance. The average fooled model for shuffle is less than crop in $0.2$.

%we start by generating adversarial examples in one iteration. We take an example of crafting adversarial examples using ResNet-18 to attack other $9$ models. We denote $5$ operations for input transformation methods, namely the crop, rotation, shuffle, scaling, and mix-up. We pair up two operations as the combination to transform the image for attacks and report the number of models fooled. We report the results in Fig. 2. It can be seen that by combining the scaling and rotation, we can achieve the maximum transferable attack success rates, indicating the optimal transformations in all possible $5 \times 5$ combinations. 

% There are two properties for the relationship between adversarial transferability and transformation operations. First, We discover that different transformations have a different impact on adversarial transferability. For example, the resize operation outperforms the other four in boosting the transferability of a white-dog image. Since there are finite ways for image transformation, there exists an optimal transformation $\varphi$ to reach the best adversarial transferability. Second, we discover that the effectiveness of one operation to boost transferability is also related to certain images. Taking figure \ref{fig:iteraion1} as an example. The optimal transformation for the rabbit image is scaling, while the optimal transformation for the image with an aircraft is rotation. 

\begin{figure}
    \centering
    \includegraphics[width=1\linewidth]{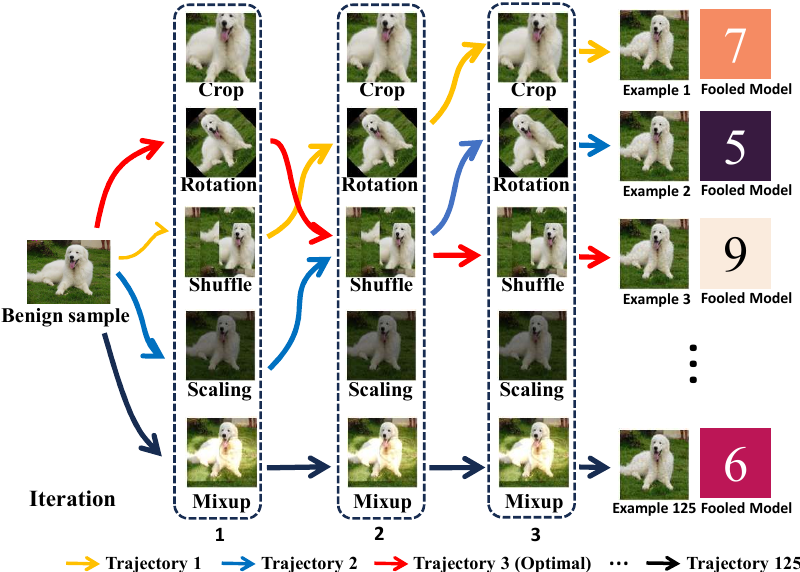}
    \caption{There exists an optimal transformation trajectory for boosting adversarial transferability. However, the search space increases exponentially with iteration number and operation number.
    }
    \label{fig:trajactory}
\end{figure}

Following the aforementioned discussion, we move on to generating adversarial examples in $3$ iterations, where we only take one operation as the image transformation to attack the image. As exemplified in  \cref{fig:trajactory}, there are $5 \times 5 \times 5$ possible trajectories to transform the image for attacks. Among these trajectories, it can achieve the best performance by first shuffling, then rotating, and last shuffling the image. It should be noted it cannot consistently achieve the best performance by increasing the number of transformations for a higher diversity. As shown in \cref{fig:trajactory}, we respectively take the scaling, shuffle, and rotation operations at each iteration in trajectory 2. However, it has the worst attack success rate among the presented results.

% First, the optimal transformation in the first iteration is possible not the optimal transformation in the second iteration. For example, as illustrated in Figure \ref{} and Figure \ref{}, the adversarial example fooled most classifiers with the transformation shuffle in the first iteration. However, scaling turns out to be the best transformation in the second iteration. Second, choosing the best transformation cannot guarantee good transferability in the second iteration. For example, the average number of fooled classifiers using shuffle (best transformation in one iteration) is smaller than the one using rotation in the first iteration. 

Generalizing the previous problem to common cases, we are motivated to identify an optimal transformation trajectory $\mathcal{T}$, which is defined as the sequence of transformation used in each iteration as $(\varphi_1, \varphi_2, \dots,\varphi_T)$, for the best adversarial transferability.  Each element $\varphi_T$ denotes the transformation used in iteration $t$. It can be formulated  as follows: 
\begin{align}
   \mathcal{T}^* = \operatorname*{argmax}_{\mathcal{T}}(\mathbb{E}[\mathcal{L}(f_{\bm{\theta}}(\bm{x}^{adv}_{\mathcal{T}}), y)]),
\label{eq:trans_opt}
\end{align}

\begin{equation}
    \mathcal{T} = (\varphi_1, \varphi_2, \dots,\varphi_T)
\end{equation}
where we denote $\bm{x}^{adv}_{\mathcal{T}}$ as the adversarial example generated by the surrogate model under transformation trajectory $\mathcal{T}$.

However, finding $\mathcal{T}^*$ is hard. First, the search space is large. For example, supposing five candidate transformations, even if we only take one operation in one iteration to transform the image,  we will still have an enormous search space for ten iterations that will be $5^{10}$. The number of possible transformation trajectories grows exponentially with increasing the number of iterations and candidate transformations. Second, we can not access the black-box model $f$, making it hard to optimize the  \cref{eq:trans_opt} directly. Besides, as identified in the previous work~\citep{yuan2022adaptive}, each image has a different optimal transformation to boost the adversarial transferability. There is no optimal transformation trajectory shared for all images.

\subsection{Methodology}

%In searching for the $\mathcal{T}^*$, we first discuss the candidates of operation $o$. we decompose some popular transformation-based attack algorithms and form a candidates pool $\Gamma$.  For example, admix \cite{} adopts two operations, mixup and scale, in boosting adversarial transferability. For mixup, admix randomly chooses three images to mix up with the benign example in a ratio of 0.2. For scale, they adopt the way proposed in SIM \cite{} and choose the number of copies as 5. Concluding from the previous, we adopt 10 categories of image operation. 1. Crop: crop the image edge with a certain range 2. Rotate: rotate the image to a given degree  3. Shuffle: cut the image into a given number of blocks and rearrange the order of these blocks. 4. Mixup: combined the benign example with random images by a given strength 5. Dropout: drop pixels from gradient calculation by a certain probability.  6. Mask: set a given block in the benign example to 0.  7. Spectrum: add  8. Scale: produce a certain number of scale copies of benign examples. 9. Resize: resize the benign example to the given size. 10. Padding: pads the benign example boundaries with zero to a given size. For more detailed implementation, please reference the supplementary. 

%To get a limited number of operations, we separate the hyperparameter space in ten identical parts and choose each division point as the hyperparameter. Thus, we have 10 operations for one operation category. We use $\{o_1, o_2, ... , o_{100}\}$ to denote the totally 100 operations and we have $\Gamma = \{o_1, o_2, ... , o_{100}\}$.

The problem of ~\cref{eq:trans_opt} can be transformed into an optimal trajectory search problem, on which reinforcement learning has shown great compatibility. We are inspired to take a reinforcement learning-based approach in solving this optimization problem to enhance adversarial transferability. 

% For one trajectory, it contains $T$ transformations. The transformation in each iteration is sampled from a probability $\bm{p}$. 

%Such a design has two advantages. First, it avoids being trapped by a large search space. Second, it obtains an efficient exploration of different operations, escaping from the multiple local optima. $\bm{p}$ is vector denotes the possibility of taking each operation. 

Supposing we have $M$ operations $\{o^1, o^2, \dots, o^M\}$ in total, the optimal transformation trajectory $\mathcal{T}$ is a temporal sequence of the combination of different operations. The probability $\bm{p}$ contains $M$ possibilities $\{p_{o^1}, p_{o^2}, ... ,p_{o^M}\}$ for each iteration. Each element $p_{o^m}$ denotes the possibility of sampling operation $o^m, m \in \{1,2,...,M\}$. And $p_{o^m}$ follows $\sum\limits_{m=1}^{M} p_{o^m} = 1$. A transformation $\varphi$ consists $K$ operations $o^k, k \in \{1,2,...,M\}$. We sampled $K$ operations from $\bm{p}$. We have the possibility of a transformation $\varphi$ by $P(\varphi) = \prod \limits_{k=1}^K p_{o^k} $.

For each iteration $t$, we sample a combination of transformation $\varphi_t$. Each transformation in $\varphi_t$ is sampled from candidates depending on $\bm{p}$. To get an optimal trajectory $\mathcal{T}=(\varphi_1,...,\varphi_T)$, we need to dynamically optimize the sampling distribution $\bm{p}$ in each iteration $t$.  We formulate the problem of searching optima $\bm{p}^*$ in each iteration as  follows,  
\begin{equation}
\begin{aligned}
    \bm{p}^* &=  \arg\max\limits_{\bm{p}} \mathbb{E}_{\varphi \sim \textbf{p}} [\mathcal{L} (f_{\bm{\theta}}(\varphi(\Tilde{\bm{x}}^{adv})), y)]  \\
    \textit{s.t.} \quad \Tilde{\bm{x}}^{adv} &= \arg\max\limits_{ \bm{x}^{adv}} \mathbb{E}_{\varphi \sim \textbf{p}}[\mathcal{L}(f_{\bm{\theta}}(\varphi(\bm{x}^{adv})), y)], \\
    %\Vert \mathbf{x}^{adv}_{\mathbf{p}} &- \mathbf{x} \Vert_{\infty} \leq \epsilon,
\end{aligned}
\label{eq:opt_p}
\end{equation}
%
%where $\mathcal{L}_{eval}$ and $\mathcal{L}_{adv}$ is the loss function for evaluating attacks and generating adversarial examples;  $\mathbf{x}^{adv} $ is the adversarial examples generated under a probability $\mathbf{p}$. 
%
which is a bi-level optimization. The inner optimization targets to optimize the adversarial example, and the outer optimization tries to find the optimal sampling probability. Following ~\cite{liu2021direct}, we adopt an one-step optimization strategy to derive the approximated $\bm{p}^*$:
\begin{equation}
    \bm{p}^* \approx \bm{p} + \rho \cdot \bm{g}_{\bm{p}},
\end{equation}
where the $\rho$ is the learning rate and $\bm{g}_{\bm{p}}$ is the gradient for $\bm{p}$. % we clarify the details in the next paragraph.  

%However, this approach still has an issue in efficiency. It brings more iteration in optimizing the outer equations for its need to sample the $\bm{x}^{adv}$ iteratively. To avoid bringing up the efficiency issue, we adopt an approximation for the inner optimization. We use the adversarial example per iteration to substitute the optimized adversarial examples. 

\begin{algorithm}
    \caption{Gradient policy for optimal augmentation search.}
    \begin{algorithmic}
        \Require Classifier $f(\cdot)$;The benign sample $\bm{x}$ with ground-truth label $y$; Loss function $\mathcal{L}(\cdot,\cdot)$; candidate operation pool $\Gamma$, the number of iterations $T$, perturbation scale $\epsilon$, policy learning rate $\rho$, number of operations $K$, number of transformations $L$, decay factor $\mu$;
    
    \State  $\alpha = \epsilon/T$, $\bm{g}_0 = 0$, $\bm{x}^{adv}_0 = \bm{x}$, $\bm{p} \sim \mathcal{N}(0,1)$
    \While{$t=1 \gets T$}
    \State 1. Under the distribution $\bm{p}$, sample $L$ transformation $\varphi_t$, each consisting of $K$ operations.

    \State 2. Transform adversarial examples: \\
     \qquad \qquad   $\varphi^l_t (x^{adv}_t) = o^{K} \oplus o^{K-1} \oplus \cdots \oplus o^1 (\bm{x}^{adv}_{t})$.
     
    \State 3. Calculate the average gradient: \\
     \qquad \qquad 
        $\bar{\bm{g}} = \frac{1}{L} \sum\limits_{l=1}^{L} \nabla_{\bm{x}^{adv}_{t-1}} \mathcal{L}(\varphi^l_t(\bm{x}^{adv}_{t-1}), y)$.

    \State 4. Update the momentum:\\
     \qquad \qquad     $\bm{g}_{t} = \mu \bm{g}_{t-1} + \frac{\bar{\bm{g}} }{|| \bar{\bm{g}}  ||_1} $.

    \State 5. Update the adversarial example: \\
      \qquad \qquad   $ \bm{x}^{adv}_t = \text{clip}(\bm{x}^{adv}_{t-1} + \alpha \cdot \text{sign}(\bm{g}_t), 0, 1) $.
     
    \State 6. Calculate the probability gradient: \\
      \qquad \qquad    $\bm{g}_{\bm{p}} = \frac{ \partial \left(\frac{1}{L} \sum\limits_{l=1}^{L} \mathbf{P}(\varphi^l_t) \mathcal{L}(f_{\bm{\theta}}(\varphi^l_t(\mathbf{x}_t^{adv})), y)] \right)}{\partial \mathbf{P}(\varphi^l_t)}$.

    \State 7. Update the probability: \\
     \qquad \qquad   $\bm{p} = \bm{p} + \rho \cdot \bm{g_{p}}$.
    \EndWhile
    \Ensure $\bm{x}^{adv}_T$.
    \end{algorithmic}
\end{algorithm}

% To keep the exploration to different operations, we use stochastic gradient ascent to optimize the probability $\bm{p}$ in each iteration $t$. 

\begin{figure}
    \centering
    \includegraphics[width=\linewidth]{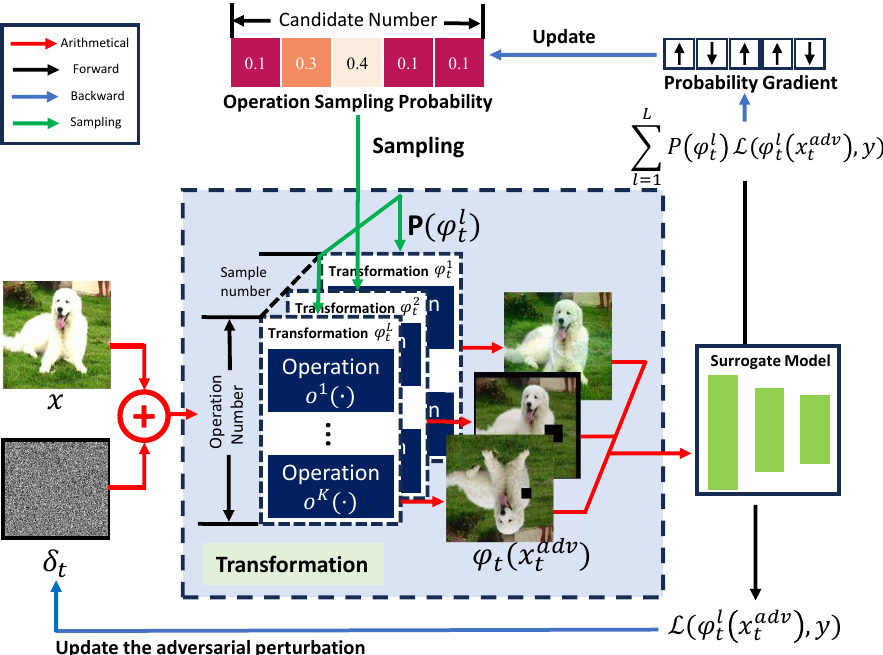}
    \caption{Overview of the pipeline in L2T. We use probability in sampling $L$ transformations and update this probability through gradient ascent.}
    \label{fig:method}
\end{figure}

\noindent \textbf{Implementation details}. We present the overview of our method in \cref{fig:method}. \underline{First}, we sample $L$ sequences of transformation $\varphi^l_t, l \in [1,2,...,L]$, depending on the sampling distribution $\bm{p}$. \underline{Next}, we get the transformed examples denoted as $\varphi^l_t(x^{adv}_{t})$. The probability of each sequence $\varphi^l_t$ is $\bm{P}(\varphi^l_t)$. We use $\varphi_t$ to denotes all $L$ transformation, $\varphi_t = \{\varphi^1_t, \varphi^2_t, ... , \varphi^L_t \}$. \underline{Then}, we use ~\cref{eq:adv_upt} to update the adversarial examples for each iteration. The gradient is calculated by loss between $L$ transformed examples and their corresponding labels. \underline{Last}, after updating the adversarial example, we recompute the approximate $\bm{p}$.  Specifically, we compute the gradient $g_{o^k}$ of each sampled operation $o^k$ as: 
\begin{equation}
\begin{aligned}
    g_{o^k} &= \frac{\partial \mathbb{E}_{\varphi_t \sim \textbf{p}}[\mathcal{L}(f_{\bm{\theta}}(\varphi_t(\mathbf{x}^{adv}_t)), y)] }{\partial \mathbf{P}(\varphi_t)} \cdot \frac{\partial \mathbf{P}(\varphi_l)}{\partial p_{o^k}} \\
    &= \frac{\partial \sum\limits_{l=1}^{L} \mathbf{P}(\varphi^l_t) \mathcal{L}(f_{\bm{\theta}}(\varphi^l_t(\mathbf{x}^{adv}_t)), y)] )}{\partial \mathbf{P}(\varphi^l_t)} \cdot \frac{\partial \mathbf{P}(\varphi_l)}{\partial p_{o^k}} \\
    & = \sum\limits_{l=1}^{L} \mathcal{L}(f_{\bm{\theta}}(\varphi^l_t(\mathbf{x}^{adv}_t), y)) \cdot \frac{\partial \mathbf{P}(\varphi^l_t)}{\partial p_{o^k}} . 
\end{aligned}
\end{equation}

We concat  the gradients for each operation as  $[g_{o^1}, g_{o^2}, \dots , g_{o^K}]$, which is denoted as $\bm{g}_{\bm{p}}$. We use gradient ascent to update $\bm{p}$ by $\bm{g}_{\bm{p}}$ with the learning rate $\rho$.

\iffalse
\begin{center}
\begin{equation}
\begin{aligned}
    \Delta_{\mathbf{\pi}} \mathcal{L}_{val}(x^{adv}_{t+1}, y) 
      &\approx  \\
    \Delta \mathbf{L}_{val} (x^{adv}_{t} + \alpha \cdot sign(\Delta_{x^{adv}_t} &\mathrm{E}_{\pi}) L_{val} (x^{adv}_{\varphi_t}, y)]). 
\end{aligned}
\end{equation}
\end{center}
\fi

%{\color{red} How to optimize the loss function? What's the update for each iteration? Introduce the alg line by line.}

\section{Experiments}
\label{sec:exp}

\subsection{Setup}
\begin{figure*}
    \centering
    \includegraphics[width=\textwidth]{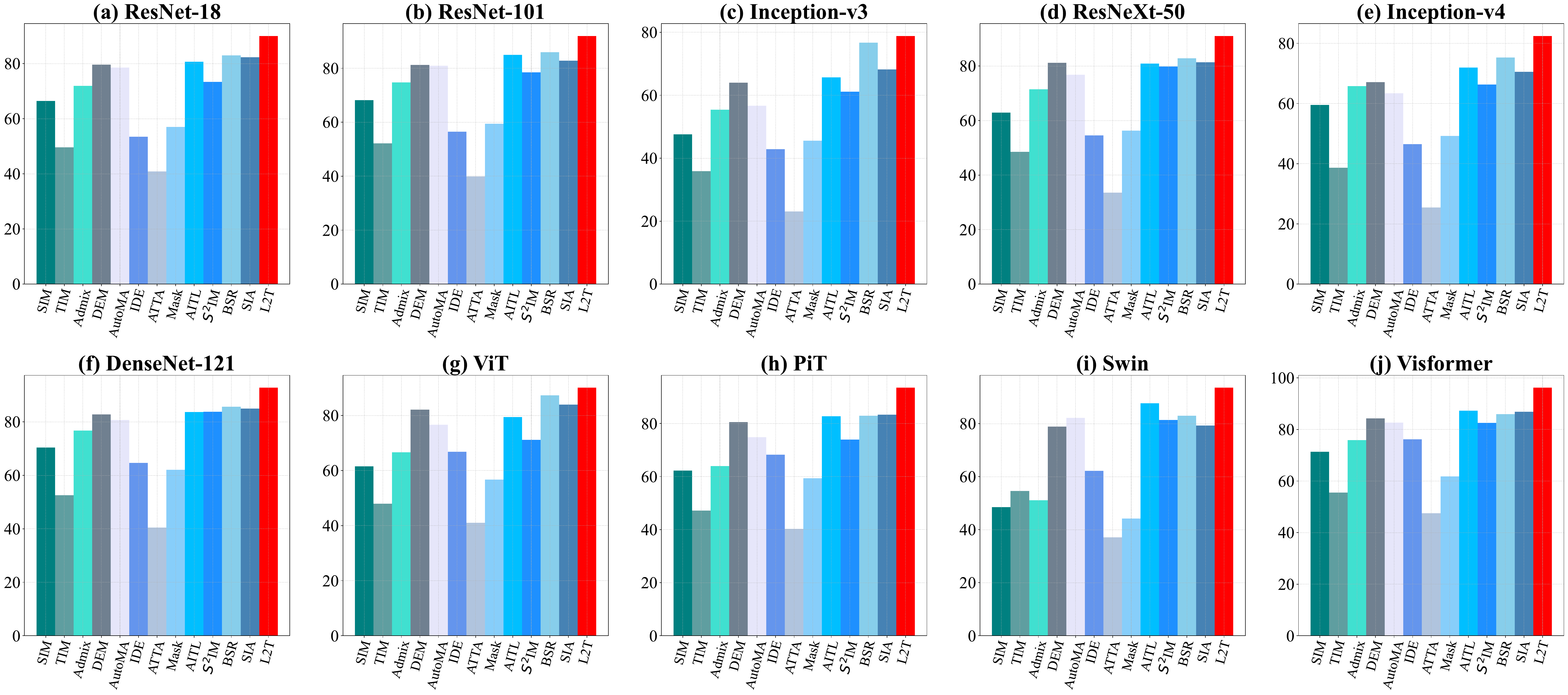}
    \caption{Average attack success rates (\%) of ten models on the adversarial examples crafted on each model. The x-axis of each sub-figure denotes different attack methods. We include the detail number in our supplementary material.}
    \label{fig:baseline}
\end{figure*}

\noindent \textbf{Models.} We evaluate the proposed method in three categories of target models. (1) Normally trained model: We select ten well-known models for experiments. ResNet-18~\cite{he2016deep}, ResNet-101~\cite{he2016deep}, ResNext-50~\cite{xie2017aggregated}, DenseNet-121~\cite{huang2017densely}, Inception-v3~\cite{szegedy2017inception}, and Inception-v4~\cite{szegedy2017inception}, ViT-B~\cite{dosovitskiy2020image}, PiT~\cite{heo2021rethinking}, Visformer~\cite{chen2021visformer}, and Swin~\cite{liu2021swin}. All of these models are pre-trained on the ImageNet dataset. (2) Adversarial trained models: we select four defense methods in our experiments. They are adversarial training (AT)~\cite{Tramr2018EnsembleAT}, high-level representation guided denoiser (HGD)~\cite{liao2018defense}, neural representation purifier (NRP)~\cite{Naseer_2020_CVPR}, and randomized smoothing (RS)~\cite{cohen2019certified}. (3) Vision API: to imitate a practical scenario, we compare the attack performance on popular vision API. We chose Google Vision, Azure AI, GPT-4V, and Bard. For categories (2) and (3), we use ensemble-based attack. We choose two CNN-based models, ResNet18 and Inception-v4, and two transformer-based models, Visformer and Swin, to construct the ensemble surrogate model. 

\noindent \textbf{Dataset.} Following previous works~\citep{ xie2019improving, wang2023structure, wang2021admix}, we randomly choose $1,000$ images from ILSVRC 2012 validation set \cite{ILSVRC15}. All images are classified correctly by the models. 

\noindent \textbf{Baseline.} We compare L2T with other input transformation adversarial methods. There are two categories of previous methods. The fixed transformation attack followed a fixed transformation scheme. We select TIM~\citep{dong2019evading}, SIM~\citep{lin2019nesterov}, Admix~\citep{wang2021admix}, DEM~\citep{zou2020improving}, IDE~\citep{xie2021improving}, Mask~\citep{fan2022maskblock}, $\rm S^2$IM~\citep{long2022frequency}, BSR~\citep{wang2023boosting}, and SIA~\citep{wang2023structure} for comparison. The learned transformation attack followed a set of transformations predicted by a trained network to generate adversarial examples. We also compare our method with learned transformation attacks, such as AutoMA~\citep{yuan2021automa}, ATTA~\citep{wu2021improving}, and AITL~\citep{yuan2022adaptive}. All these methods are integrated with MI-FGSM~\citep{dong2018boosting} to generate adversarial examples. 

\noindent\textbf{Evaluation Settings.}
We follow the hyper-parameter setting of MI-FGSM and set the perturbation budget $\epsilon = 16$, number of iteration $T = 10$, step size $\alpha = \epsilon/T  = 1.6$ and decay factor $\mu = 1$. For our method, we adopt the number of operations as $2$, the number of samples as $10$, and the learning rate $\rho$ as $0.01$. For the candidate operation, we chose ten categories of transformations. Each category contains ten specific operations with different parameters. We will discuss the detailed settings of our method and other baselines in the supplementary materials. 
%For the settings of other baseline methods, please refer to the supplementary materials as well. 
\begin{figure}
    \centering
    \includegraphics[width=0.9\linewidth]{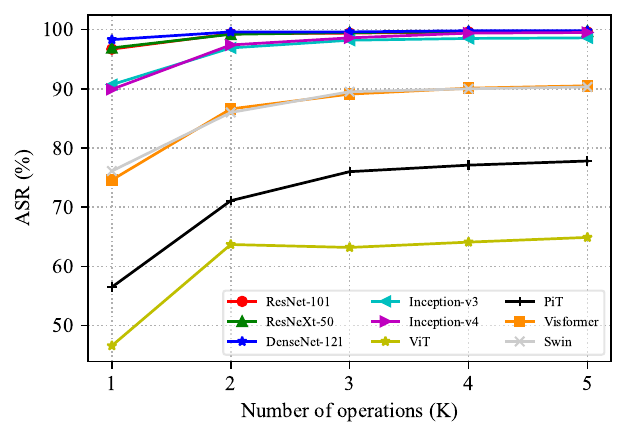}
    \caption{Attack success rates (ASR) (\%) of adversarial examples generated by L2T with various number of operations $K$. We include the detail number in our supplementary material. }
    \label{fig:ops}
\end{figure}

%For DIM, we adopts the transformation probability of 0.5 and the resize rate as 1.1. We choose the number of copies $m=5$, for SIM. For DEM, the resize ratio is set to [1.14, 1.27, 1.4, 1.53, 1.66].  For SSM, we adopt the number of spectrum as 20 and $\rho$ as 0.5.  Admix randomly mixes 3 images from other categories with the strength of 0.2 and 5 scaled images for each admixed image. For MaskBlock, the patch size is setted to 56. 
\begin{figure*}
    \centering
    \includegraphics[width=\textwidth]{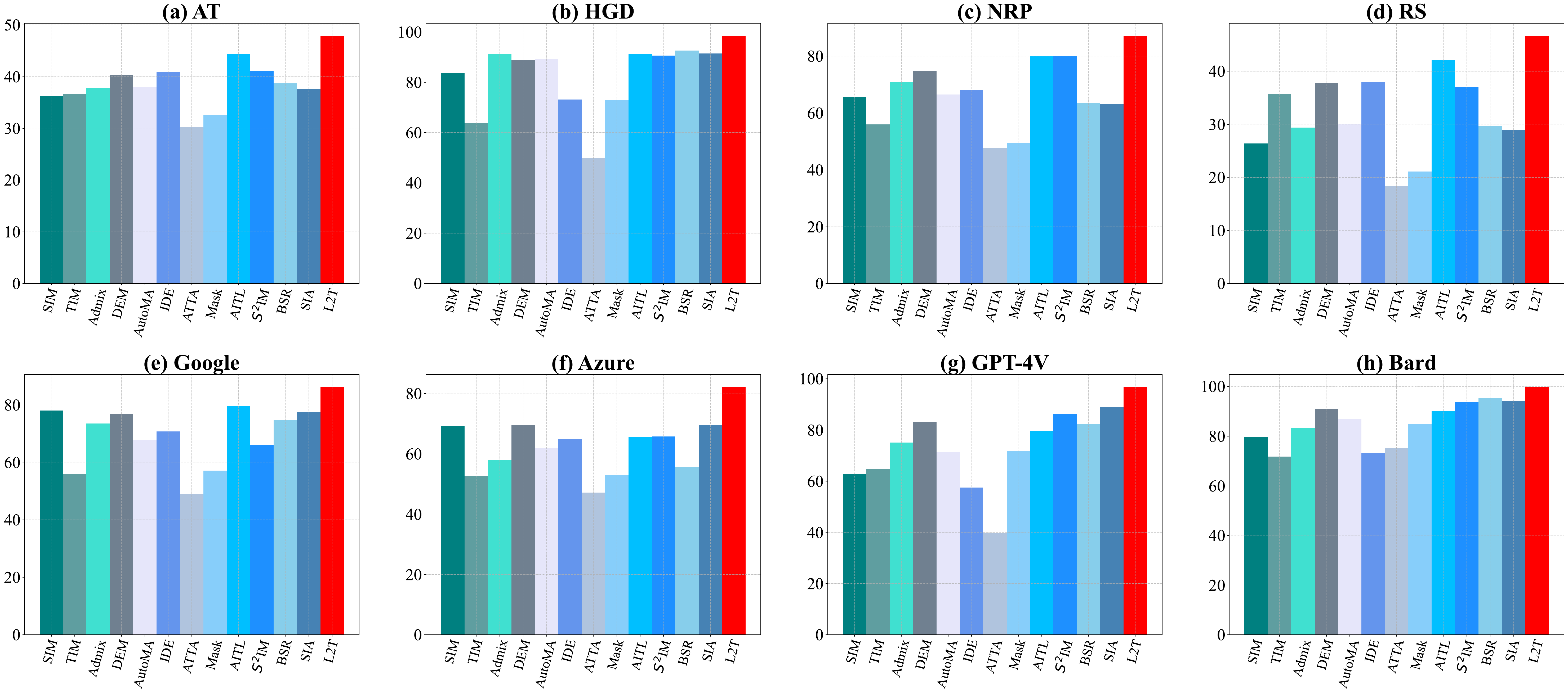}
    \caption{We integrate the ensemble-based attack with input transformation and evaluate the performance on defense methods and popular vision APIs. We include the detail number in our supplementary material.}
    \label{fig:defense}
\end{figure*}

% \subsection{Evaluations on separate iterations}

% We discussed the transferability between each iterations. From the Figure ?, we break down the attack success rate (\%) into iterations. For single iteration, our methods achieves a better transferablity (\%) by finding an better input transformation. 

%add conclusion for the evaluation 

\subsection{Evaluations on single models}

%Our proposed L2T consistently outperforms various input transformation based attacks. To evaluate the transferability of adversarial examples from different methods, we use the attack success rate (ASR) to indicate the adversarial transferability. We compare L2T with ten fixed transformations and three learning-based transformation methods. We generate the adversarial examples on a single model and test them on all models. 

Our proposed L2T exhibits better adversarial transferability to various input transformation based attacks. We take a single model as the surrogate model and evaluate the average attack success rate (ASR), \ie, the average misclassification rates across ten models. We summarized our results in Figure~\ref{fig:baseline}. Each subfigure denotes the attacker generates the adversarial examples on the corresponding models and its x-axis denotes the attack algorithm used.

%Figure~\ref{fig:baseline} shows the average ASR on ten models. First, we observe that L2T outperforms all other attackers, regardless of the surrogate model. It means that L2T is capable of all surrogate models with dynamic transformation. For other baseline methods, the performance on different surrogate models varies. For example, the BSR performs to be the strongest baseline on ResNet-18. However, the BSR cannot remain efficient when the surrogate model is changed to Swin or PiT. These results also strengthen our argument that we should dynamically choose the transformation. Specifically, L2T outperforms the other baseline by 22.9\% on average and achieves a better attack success rate than the best baseline in at least 2.9\% among all test cases. 

First, we observe that L2T consistently outperforms all other attackers, regardless of the surrogate model. Other baseline methods have various adversarial transferability according to the surrogate models. For example, the BSR performs to be the strongest baseline on ResNet-18. However, the BSR cannot remain efficient when the surrogate model is changed to Swin or PiT. In contrast, our proposed L2T is suitable for all the surrogate models being tested. These results also strengthen our argument that we should dynamically choose the transformation to fit the surrogate models. Specifically,  in the worst case (subfig. c), our proposed L2T still outperforms the strongest baseline ($S^2$IM) by $2.1\%$. Overall, L2T outperforms the other baseline by 22.9\% on average ASR.

\subsection{Evaluations on defense methods}
\label{sec:exp_def}
L2T is also capable of adversarial robust mechanisms. We test the attack performance of L2T against several defense mechanisms, including AT, HGD, NRP, and RS. We choose the ensemble setting to attack these defense approaches. We use the ensemble of four models, ResNet-18, Inception-v4, Visformer, and Swin, as the surrogate model. We summarized our results in Figure~\ref{fig:defense} (a), (b), (c), and (d). Each subfigure denotes the model to be attacked and its x-axis denotes the attack algorithm used.

%Many defense methods~\cite{cohen2019certified, Tramr2018EnsembleAT} have been proposed to defend against adversarial attacks. We conducted experiments to study the effectiveness of these defenses. We chose four networks, ResNet-18, Inception-v4, Visformer, and Swin, to perform ensemble-based attacks. We adopt the adversarial examples generated on these four models simultaneously to attack the defense models. We test the performance of different attackers against four defense methods.

From Fig.~\ref{fig:defense}, it is clear that L2T remains efficient. L2T consistently outperforms other methods against various defense methods. Notably, it achieves the attack success rate of $47.9\%$, $98.5\%$, $87.2\%$, and $46.7\%$ on AT, HGD, NRP, and RS, respectively. Even on the certified defense RS, the strongest defense among the four, L2T achieves the attack success rate of $46.7\%$, which exceeds the best baseline (AITL) by $4.6\%$. This is also the biggest improvement L2T made compared to other defenses. This indicates that the dynamic of iteration also exists in the adversarial robust mechanism, which can be used to dimish the its performance.

%Although the RS method presents to be the strongest defense method, our method could still exceed the best baseline by $4.6\%$, which means a $10\%$ improvement. Dynamical transformations can help adversarial examples increase the adversarial transferability of defense methods. This also sheds light on developing new defense methods.

% The vision API can be separated into two categories. First, these APIs target only vision tasks. An image is sent to the server, and the user gets the predicted class label as a return // Second, these APIs incorporate multi-modal large language models (MLLMs). The user provides an image and a prompt to get the class label.

\subsection{Evaluations on vision API}
Our proposed L2T can also perform well in realistic scenarios. To imitate the real-world application, we test the performance of L2T on Vision API. We use the same setting in sec.~\ref{sec:exp_def} to craft adversarial examples. We choose Google Vision (Figure~\ref{fig:defense} (e)) and Azure AI (Figure~\ref{fig:defense} (f)) to evaluate attacks on vision-only API.  We also choose ChatGPT-4V (Figure~\ref{fig:defense} (g)) and Gemini (Figure~\ref{fig:defense} (f)) to evaluate attacks on the foundation model API.

\begin{table*}[]
\caption{Attack success rates (\%) of adversarial examples by L2T  and Rand (randomly choose transformation in each iteration).}
\resizebox{\linewidth}{!}{
\begin{tabular}{ccccccccccc}
\hline
ResNet-18 & ResNet-101 & ResNeXt-50 & DenseNet-121 & Inception-v3 & Inception-v4 & ViT   & PiT  & Visformer & swin    \\ \hline
Rand     & 52.35     & 59.06     & 53.19       & 56.64        & 43.01        & 44.41 & 58.41 & 54.48     & 65.08 \\
\rowcolor{mygray} L2T (Ours)       & 90.00        & 91.90      & 91.00          & 92.80         & 78.80         & 82.40  & 90.10  & 93.50      & 96.20 \\ \hline
\end{tabular}}

\label{tabs:rand}
\end{table*}

%From Fig.~\ref{fig:defense}, L2T performs the best attack performance consistently under real-world scenarios. For example,  in Azure's case, we achieve a 12.1\% margin over the best baseline. For LLMs vision API, our methods can achieve nearly 100\% attack success rate. These results show the vulnerability of current vision applications. 

As shown in Fig.~\ref{fig:defense}, L2T is generally the best attacker to the real-world API. All attacks perform better on foundation model API than vision-only API. For vision-only API, L2T outperforms the strongest baseline by $8.7\%$ and $12.6\%$, respectively. For foundation model API, L2T achieves nearly $100\%$ attack success rate on both GPT-4V and Gemini.

\begin{figure}
    \centering
    \includegraphics[width=0.9\linewidth]{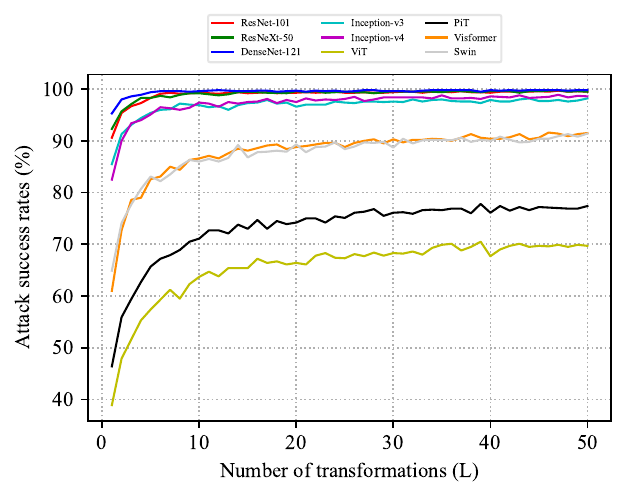}
    \caption{Attack success rates (ASR) (\%) of adversarial examples generated by L2T with various number of transformations $L$. We include the detail number in our supplementary material.}
    \label{fig:copies}
\end{figure}

%\subsection{Evaluations on Google's Vision API}

%\subsection{Evaluations on targeted attacks}

\subsection{Ablation study}

%The number of operations determines how many operations are used to transform the input image before it is sent to classifiers  the difference between 2 operations and more is marginal  to avoid breaking the structure of the image and introducing diversity for the gradient. 

\noindent \textbf{On the numbers of operation $K$.}
As shown in \cref{fig:ops}, we study the impact of  $K$ on adversarial transferability. We craft the adversarial example on ResNet-18 and evaluate them on the other nine models. There is a clear difference between one operation and two operations. The average attack success rate increases by $8.09\%$, from $80.89\%$ to $88.98\%$. However, when the $K \geq 3$, the improvement becomes marginal. The average attack success rate only increases by $2.29\%$ when  $K$ is increased from $2$ to $5$. Thus,  $K$ should be moderately settled as $2$.

\begin{figure}
    \centering
    \includegraphics[width=0.9\linewidth]{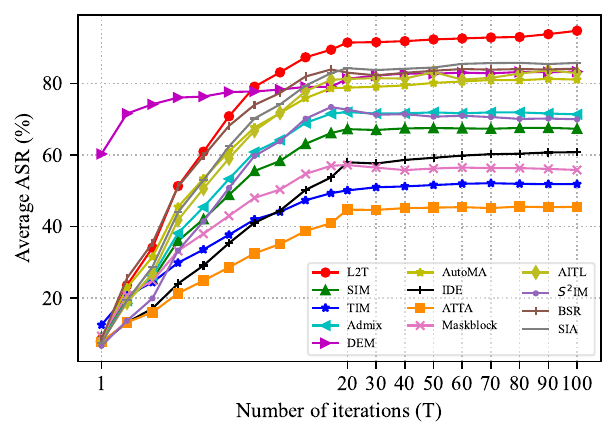}
    \caption{Average attack success rates (ASR) (\%) of adversarial examples generated by L2T with various number of steps $T$. We include the detail number in our supplementary material.}
    \label{fig:steps}
\end{figure}

\noindent  \textbf{On the number of transformations $L$.}
We conducted experiments on the number of transformations $L$. We craft the adversarial example on ResNet-18 and evaluate them on the other nine models. We choose $L$ from 1 to 50. From Fig.~\ref{fig:copies}, we observe that the adversarial transferability improves steadily with the number of transformations. The increase is significant when the number of transformations grows from 1 to 20, which improves from an average attack success rate of $75.7\%$ to an average attack success rate of $91.1\%$. However, transferability does not increase significantly after the number exceeds $20$, where the average attack success rate only increases $1.5\%$. To keep the balance between computation efficiency and adversarial transferability, we suggest the number of samples set to 20.

%\input{tabs/num_copy}

% \textbf{On the number of learning rate} 0:random, fixed learning rate, learning rate with decay strate

\noindent  \textbf{On the number of iterations $T$.}
We discuss the number of iterations among different attack approaches. We craft the adversarial example on ResNet-18 and compare the average attack success rate of 10 models. As shown in Fig.~\ref{fig:steps}, for all the attack methods, the attack success rate increased steadily for the first $10$ iterations. L2T achieves the fastest speed of increase, which reaches $89.47\%$ at iteration $10$. After $10$ iterations, most of the methods struggled to make improvements. For example, the Admix goes around $71\%$. The performance of $\rm S^2$IM even decreases from $73\%$ to $70\%$. Meanwhile, L2T still maintains a stable increase, from $89.47\%$ to $94.77\%$. 

%data should be a lot since too much figures
\begin{figure}
    \centering
    \includegraphics[scale=0.195]{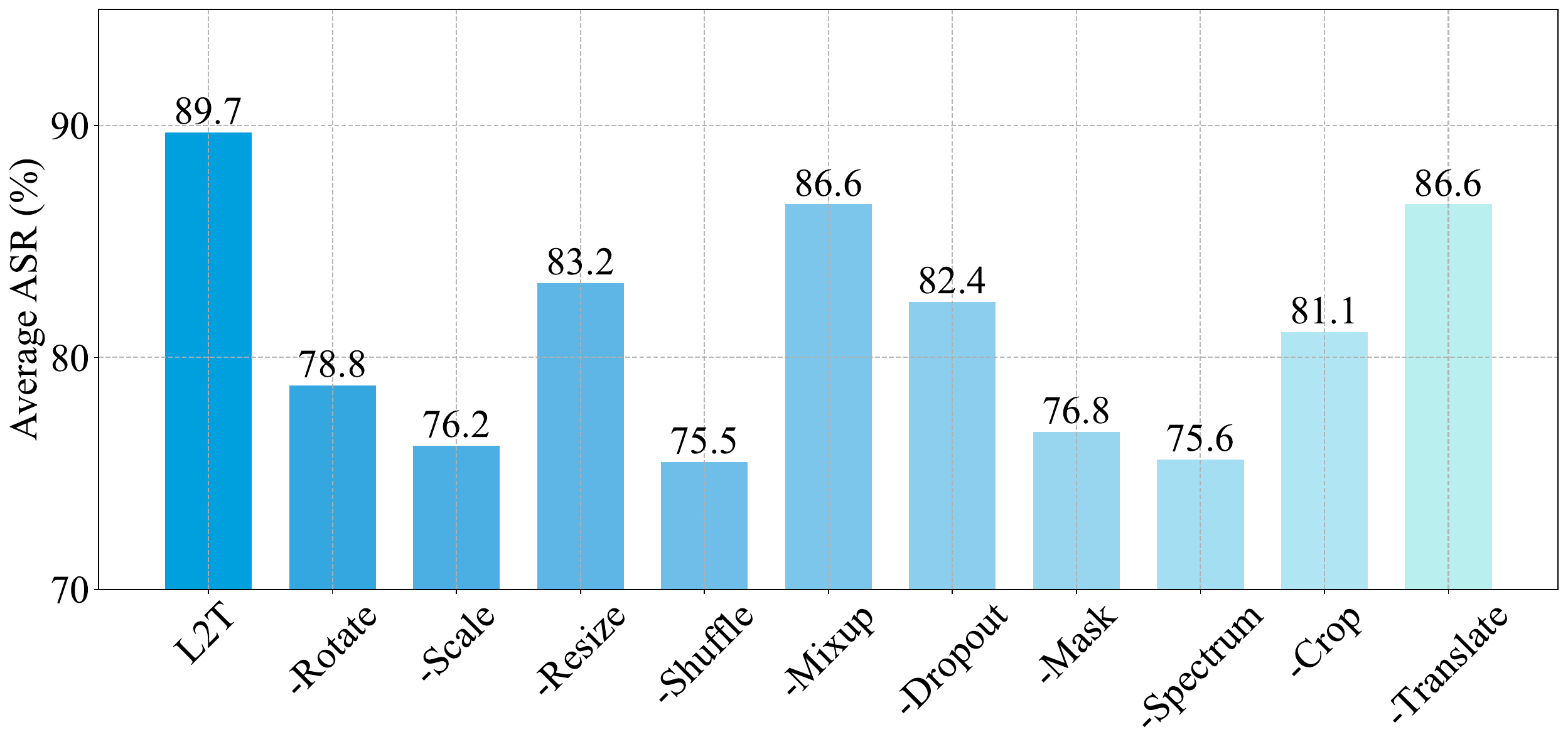}
    \caption{The average attack success rates (\%) of adversarial examples crafted by L2T and L2T without a single transformation. $-$ indicates removing such transformation.}
    \label{fig:abl}
\end{figure}

\noindent \textbf{Comparison with random sampling.} We compare the learnable strategy with random sampling. As shown in \cref{tabs:rand}, there is a clear gap of the attack success rate between random sampling and gradient-guided sampling. The minimum difference is $31.12\%$ with setting  Visformer as the surrogate model. For other surrogate models, the gap is even larger. This experiment indicates random sampling cannot effectively sample the best transformation trajectory, and the transformation in each iteration needs to be chosen carefully. 

\noindent \textbf{Operation candidates analysis.}
We conducted an ablation study for the operation candidates. We subtract each operation in the candidates and conduct L2T on the updated operation candidates. From Fig.~\ref{fig:abl}, we observe that subtracting any operations will lead to a performance decrease. For example, by subtracting the scale operation, the performance decreases for $23.5\%$. Meanwhile, subtracting mixup and translation only results in a $3.1\%$ decrease.

% \textbf{On the number of i}

% \textbf{combine with Gradient-based attack and model-based attack}

\section{Conclusion}
\label{sec:conclusion}

In this paper, we study the dynamic property for input transformation. Utilizing this property, we propose L2T to optimize the input transformation in each iteration. By updating a sampling probability, our method provides an approximate solution to input transformation optimization. Our experiments further study the effectiveness of our methods. Our method performs consistently well among different targeted models. This paper provides a new perspective to understand the transferability of adversarial examples. 

\noindent \textbf{Acknowledgement}. This work was supported by NSF under grant 2202124 and the Center of Excellence in Data Science, an Empire State Development-designated Center of Excellence. The content of the information does not necessarily reflect the position of the Government, and no official endorsement should be inferred.
{
    \small
    \bibliographystyle{ieeenat_fullname}
    \bibliography{main}
}

\clearpage
\onecolumn
\appendix
\renewcommand{\appendixpagename}{Appendix}
\appendixpage

%\clearpage
%\setcounter{page}{1}
%\maketitlesupplementary

\section{Experiments Settings}
\subsection{Baseline methods}

\begin{itemize}

\item \textbf{TIM:}
TIM adopts a translation operation that shifts the benign example by $i$ and $i$ pixels along the two dimensions, respectively. TIM uses a kernel matrix in gradient calculation to replace the translation. In our experiments, we chose the Gaussian kernel as $\tilde{W}_{i,j} = \frac{1}{2\pi\sigma^2} \exp\left(-\frac{i^2 + j^2}{2\sigma^2}\right) $ and $W_{i,j} = \frac{\tilde{W}_{i,j}}{\sum_{i,j} \tilde{W}_{i,j}}$.

\item \textbf{SIM:}
The scale-invariant method (SIM) scales every pixel by a set of levels and uses these scaled images for gradient calculation. In our experiments, we choose the number of scale samples $m = 5$ and the scale factor $\gamma_i = 1/2^i$.

\item \textbf{Admix:}
Admix randomly mixes the benign examples with images from other categories and scales the mixed examples in different scales. We set the scale copies $m_1 = 5$ and scale factor $\gamma_i = 1/2^i$ and random sample images $m_2 = 3$ and mixup strength as $0.2$.

\item \textbf{DEM:}
DEM provided an ensemble version of diversity invariant methods, which uses five transformed copies for gradient calculation. In our experiments, we set the diversity list to [340, 380, 420, 460, 500]. 

\item \textbf{Masked:}
Maskblock separates the images into several blocks and sequentially masks every block in the benign examples. Thus, the number of transformed copies is equal to the number of blocks. We set the number of blocks to 16 in our experiments.

\item \textbf{IDE:}
IDE conducts input dropout on a being example at different rates and gets multiple transformed examples to form an ensemble attack. In our experiments, we choose the dropout rate to be 0.0, 0.1, 0.2, 0.3, 0.4, and the weight factor as equal.

\item \textbf{$\rm{S}^2$IM:}
$\rm{S}^2$IM provides a frequency domain perspective of input transformation, which utilizes DCT and IDCT techniques in transformation. In our experiments, we set the tuning factor $\rho$ = 0.5 and the standard deviation $\sigma$ the same with perturbation scale $\epsilon$ and the number of spectrum transformations $N = 20$.

\item \textbf{BSR:}
BSR splits the input image into several blocks and then randomly shuffles and rotates these blocks. In our experiments, we split the image into $2x2$ blocks with the maximum rotation angle $24\%$ and calculate the gradients on $N = 20$ transformed images.

\item \textbf{SIA:}
SIA decomposed the images into several blocks and transformed each block with an input transformation choosing from seven transformation candidates \footnote{Vertical Shift, Horizontal Shift, Vertical Flip, Horizontal Flip, Rotate, Scale, Add noise, Resize, DCT, Dropout}. We followed the suggested settings in the paper and chose splitting number $s = 3$, number of transformed images for gradient calculation $N = 20$.

\item \textbf{AutoMA:}
AutoMA targeted finding a strong model augmentation policy to boost adversarial transferability. Following the setting in the paper, we trained the augmentation policy search network on 1000 images from ImageNet \cite{ILSVRC15} validation set, which does not overlap with the benign example set. We adopt the transformation number $m=5$ and set the ten operation types and their corresponding magnitude the same as the original paper. %\footnote{1.Type: Resize\&Padding; Magnitude:[299,1]. 2.Type: Translation; Magnitude:[0,299]. 3.Type: Rotation; Magnitude:[0,90]. 4.Type:Horizontal Flip; Magnitude:None }

\item \textbf{ATTA:}
ATTA uses a two-layer network to mimic the transformation function. The benign examples are first passed through this transformation network and then sent for calculating the adversarial perturbations. We use the data from ImageNet \cite{ILSVRC15} training partition to train the transformation network. We trained different transformation networks according to the surrogate models. For the training hyperparameters, we follow the settings from the authors. 

\item \textbf{AITL:}
AITL introduces selecting input transformations by different benign examples. AITL trains three networks to predict the input transformations for every image.  We adopt the 20 image transformations in the same paper and use the pre-train model weights from the authors to initialize the above networks. We set the number of iterations during optimizing the image transformation feature to $1$, the corresponding step size to $15$, and the number of image transformation operations to $4$.
\end{itemize}

\subsection{Learning to Transform}
We decomposed the existing methods and concluded their input transformation methods. We formulate the transformation candidates in 10 categories. 
\begin{itemize}
    \item \textbf{(1) Rotate}: Rotate refers to turning the image around a fixed point, usually its center, by a certain angle. The domain of angle is $[0, 360]$. We choose 10 angles from the domain, and the interval between the two angles is identical. Thus, we form 10 operations for the rotate category. The smallest rotation angle is $36^\circ$, and the biggest rotation angle is $360^\circ$.

    \item \textbf{(2) Scale}: the scale category comes from $\textbf{SIM}$. we form 10 operations in our experiments. Each operation differs in scale factor $\gamma = 1/2^i, i \in [1,2,...,10]$.

    \item \textbf{(3) Resize}: Resize refers to removing the margin part of examples and resizing the main body of the benign examples. We chose 10 resize rates for our experiments, which are $0$,$0.1$, $0.2$, $0.3$, $0.4$, $0.5$, $0.6$, $0.7$, $0.8$, and $0.9$ respectively. 

    \item \textbf{(4) Pad}: the pad category comes from $\textbf{DIM}$. We choose to pad the bengin examples to different sizes where the size of the padded example will be $[size \times size]$. We chose 10 different sizes, which are 246.5, 257.6, 268.8, 280.0, 291.2, 302.4, 313.6, 324.8, 336.0, and 347.2.

    \item \textbf{(5) Mask}: The mask category comes from Masked, which separates the examples into several blocks and randomly blocks one of the blocks. We control the number of blocks and choose 4,9,16,25,36,49,64,81,100,121 in specific. 

    \item \textbf{(6) Translate}: the translated category comes from $\textbf{TIM}$. We shift the benign examples into 10 levels, which are 10pixel, 20pixel, 30pixel, 40pixel,50pixel, 60pixel, 70pixel, 80pixel,
90pixel, 100pixel, along the x-axis and y-axis. 

    \item \textbf{(8) Shuffle}: The shuffle category comes from \textbf{BSR}, which separates the examples into several blocks and randomly reorders these blocks. We control the number of blocks and choose 4,9,16,25,36,49,64,81,100,121 in specific.

    \item \textbf{(9) Spectrum}: the spectrum category comes from \textbf{$\rm{S}^2$IM}, which adds noise in the spectrum domain of benign examples determined by strength $\rho$. We set ten different $\rho$ as 0.1, 0.2, 0.3, 0.4, 0.5, 0.6, 0.7, 0.8, 0.9, 1.0.

    \item \textbf{(10) Mixup}: the mixup category comes from \textbf{Admix}. We choose two mixup strengths, 0.2 and 0.4, and five mixup numbers as 1, 2, 3, 4, 5. Thus, we form 10 operations by combining the two settings.
\end{itemize}

\section{Numerical Results}

\noindent \textbf{Comparison with advanced methods}: 
We include detailed results of the comparison with different baselines in \cref{tabs:resnet18}, \cref{tabs:resnet101}, \cref{tabs:dense121}, \cref{tabs:next50}, \cref{tabs:Inc-v4}, \cref{tabs:inc-v3}, \cref{tabs:vit}, \cref{tabs:pit}, \cref{tabs:vis}, \cref{tabs:swin}. For each table, we choose one model from ten models as the surrogate model and use the adversarial examples to attack all these ten models. 

We show the attack success rate on adversarial examples crafted on ten different models corresponding to \cref{fig:baseline}. \cref{tabs:resnet18} is the detailed results for \cref{fig:baseline}(a). \cref{tabs:resnet101} is the detailed results for \cref{fig:baseline}(b). \cref{tabs:inc-v3} is the detailed results for \cref{fig:baseline}(c). \cref{tabs:next50} is the detailed results for \cref{fig:baseline}(d). \cref{tabs:Inc-v4} is the detailed results for \cref{fig:baseline}(e). \cref{tabs:dense121} is the detailed results for \cref{fig:baseline}(f). \cref{tabs:vit} is the detailed results for \cref{fig:baseline}(g). \cref{tabs:pit} is the detailed results for \cref{fig:baseline}(h). \cref{tabs:swin} is the detailed results for \cref{fig:baseline}(i). \cref{tabs:vis} is the detailed results for \cref{fig:baseline}(j). The effectiveness of each attack varies significantly across different models. The L2T attack shows remarkably high effectiveness across all models, which outperforms all the other methods on all ten models.

\noindent \textbf{Evaluation on the defense methods and cloud APIs}: 
We include the detailed results across different defense methods and vision API in \cref{tabs:defense&API} corresponding to \cref{fig:defense}. The L2T attack, highlighted in gray, shows exceptionally high success rates across almost all defense methods and APIs, particularly against Bard and GPT-4V.  

\noindent \textbf{Ablation study on the number of iterations}: 
We include the detailed results on the different iterations in \cref{tabs:iteration} corresponding to \cref{fig:steps}. For most attacks, success rates increase as the number of iterations increases. This indicates that more iterations generally lead to more effective adversarial examples. After a certain number of iterations (around 20-30 for many attacks), the increase in success rate slows down or plateaus. For example, the L2T attack's success rate increases significantly up to about 30 iterations and then grows more slowly.

%\subsection{Detailed results for sample number study}
\noindent \textbf{Ablation study on the number of samples}: 
We include the detailed results on the different iterations in \cref{tabs:samples} corresponding to \cref{fig:copies}. This suggests that using more samples to generate adversarial examples can lead to more effective attacks.

\noindent \textbf{Ablation study on the number of operations}: 
We include the detailed results on the different iterations in \cref{tabs:ops_num} corresponding to \cref{fig:ops}. As the number of operations increases, there is a general trend of increasing success rates across most models. However, the increase is not significant after the number of operations exceeds $2$.

\section{Examples on attacking the Multi-modal Large Language Models}

\begin{figure*}[hbp]
    \centering
    \includegraphics[width=\linewidth]{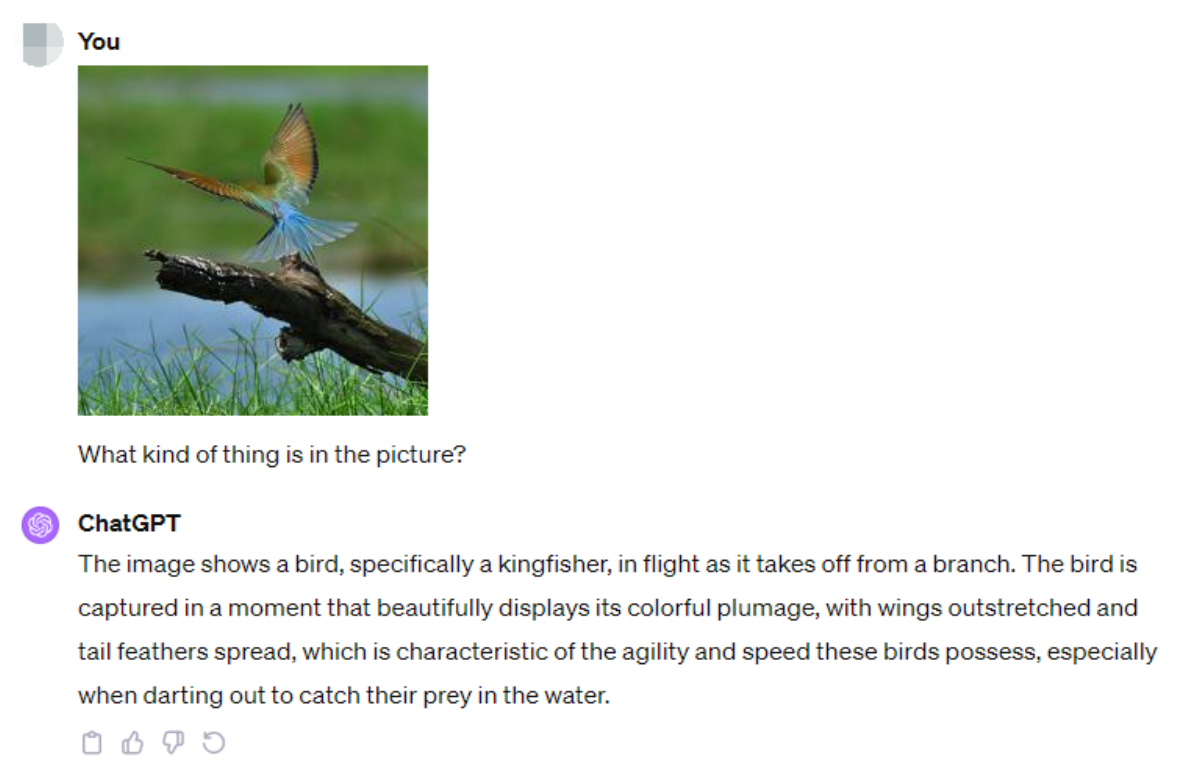}
    \caption{The conversation with ChatGPT for the benign example}
    \label{fig:gpt_clean}
\end{figure*}

\begin{figure*}[hbp]
    \centering
    \includegraphics[width=\linewidth]{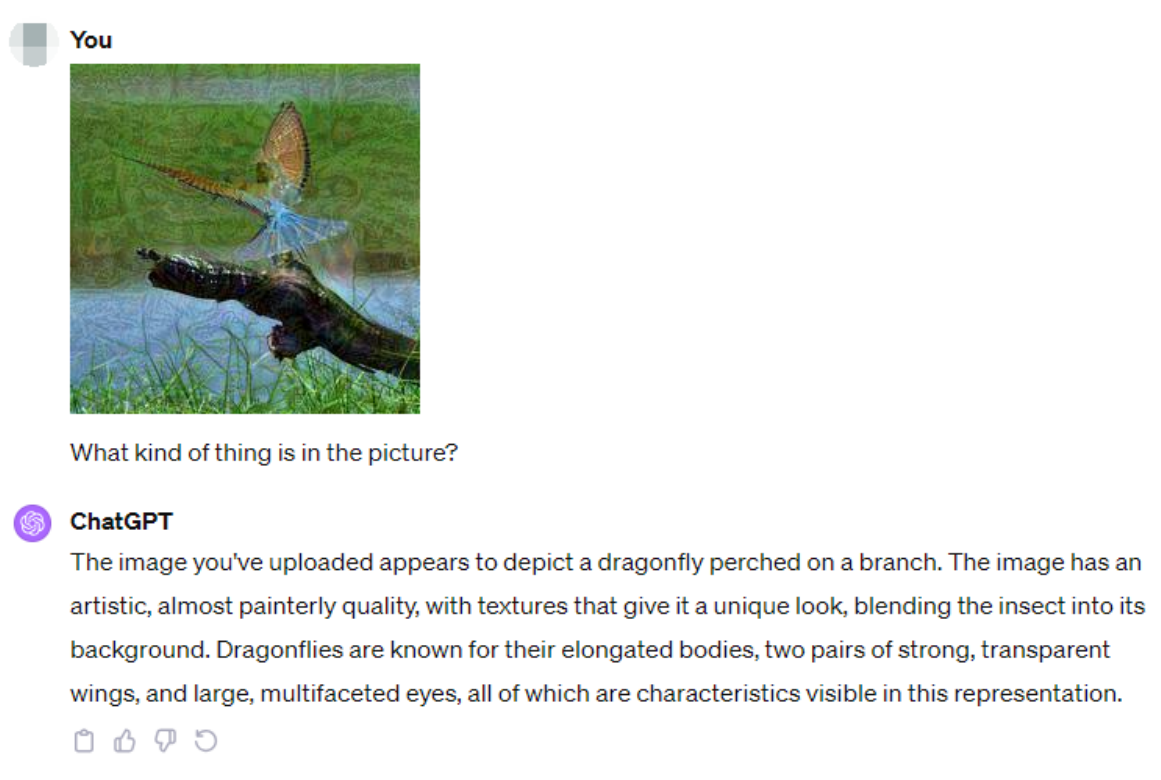}
    \caption{The conversation with ChatGPT for the adversarial example}
    \label{fig:gpt_noise}
\end{figure*}

\begin{figure*}
    \centering
    \includegraphics[width=\linewidth]{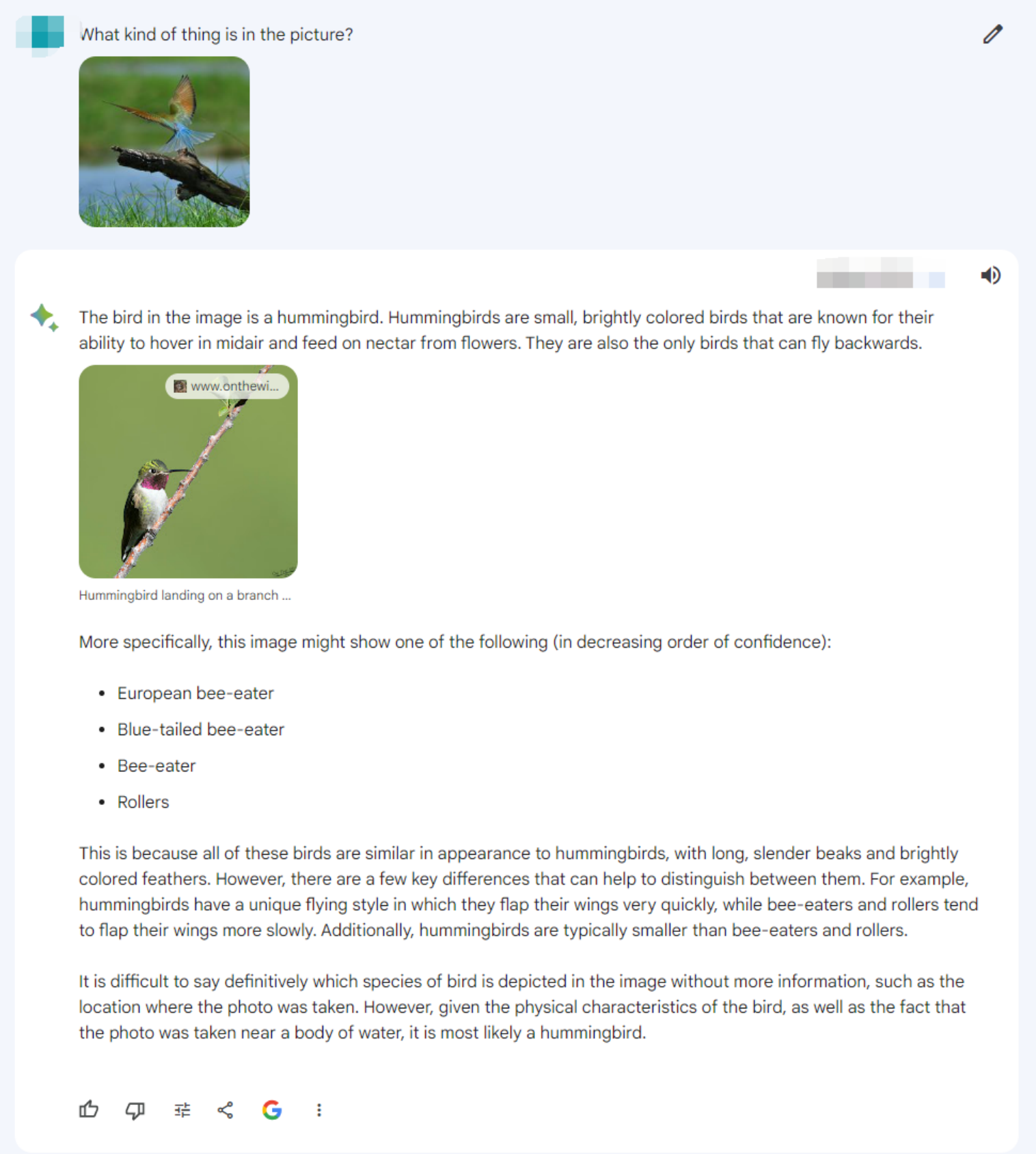}
    \caption{The conversation with Bard for the benign example}
    \label{fig:bard_clean}
\end{figure*}

\begin{figure*}
    \centering
    \includegraphics[width=\linewidth]{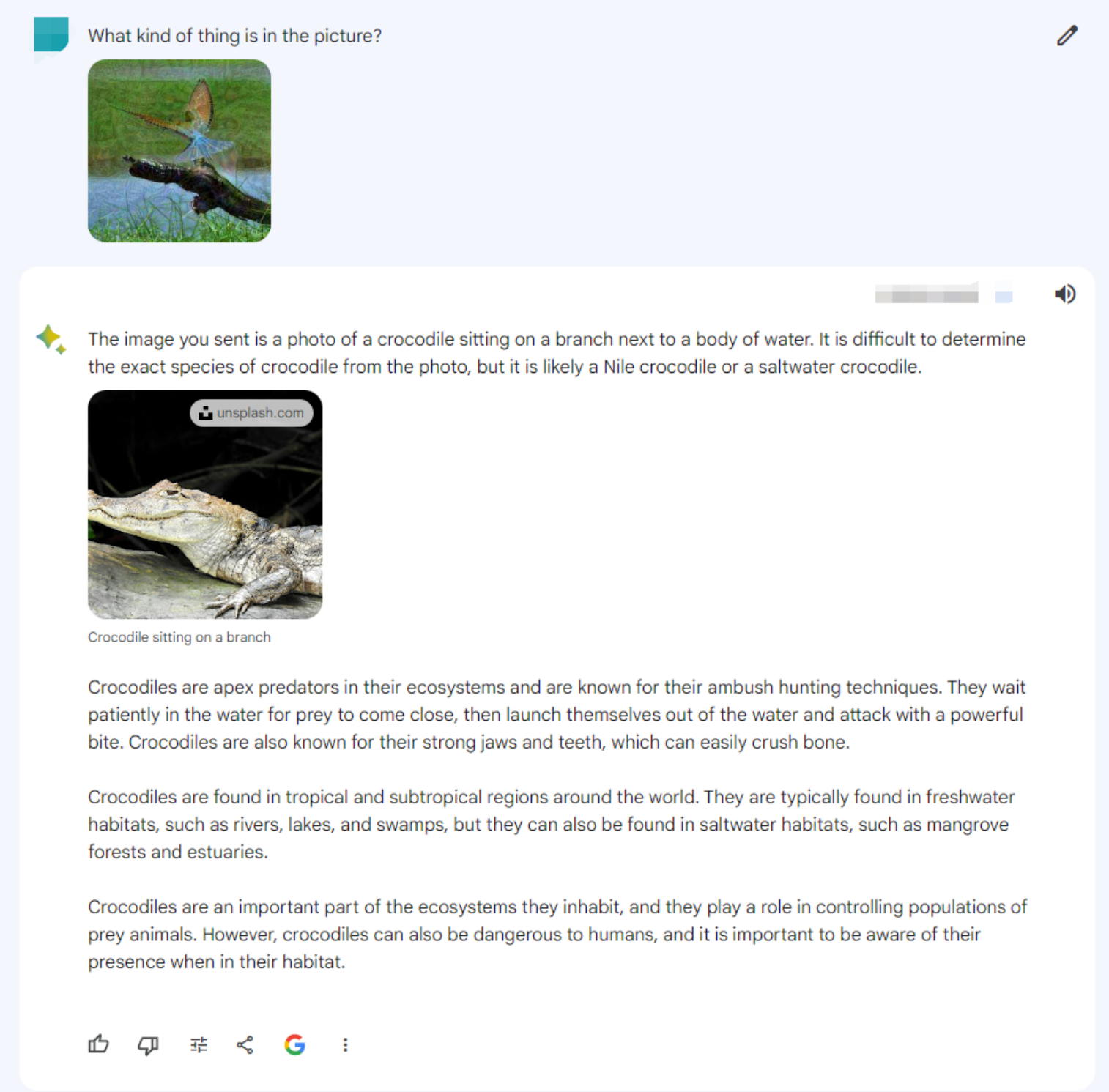}
    \caption{The conversation with Bard for the adversarial example}
    \label{fig:bard_noise}
\end{figure*}

To show the scalability of L2T, we also conducted experiments on multi-modal large language models (MLLMs). As shown in \cref{fig:bard_clean}\cref{fig:gpt_clean}, both GPT-4V and Bard can classify the benign example correctly into the ``bee-eater''. We use L2T to generate the adversarial examples against ResNet-18. As shown in \cref{fig:bard_noise}\cref{fig:gpt_noise}, the Bard classified the adversarial example as a crocodile, and GPT-4V classified it as a dragonfly. It shows the vulnerability of MLLMs, posing great challenges in developing robust MLLMs.

%We show the attack success rate on adversarial examples crafted on ResNet-18 in \cref{tabs:resnet18}. The effectiveness of each attack varies significantly across different models. For instance, the I-FGSM attack has a much lower success rate on models like ViT and PiT compared to ResNet-18. The L2T attack shows remarkably high effectiveness across all models, with the lowest success rate being 63.7\% on ViT and the highest on the surrogate model (ResNet-18).

%We show the attack success rate on adversarial examples crafted on ResNet-101 in \cref{tabs:resnet101}. The average success rate column helps compare the overall effectiveness of each attack. Attacks like BSR, AITL, and L2T show high average success rates, while L2T exhibits the best performance at 91.9\%, indicating their general effectiveness across different models.

%We show the attack success rate on adversarial examples crafted on DenseNet-121 in \cref{tabs:dense121}. 

\begin{table*}[b]
\setlength\tabcolsep{3pt}
\caption{Attack success rate (\%) across ten models on the adversarial examples crafted on ResNet-18 by different attack}
\resizebox{\textwidth}{!}{
\begin{tabular}{cccccccccccc}
\toprule
Attack       & Res-18 & Res-101 & NeXt-50 & Dense-121 & Inc-v3 & Inc-v4 & ViT  & PiT  & Visformer & Swin & Average \\ 
\midrule
I-FGSM       & 100.0    & 30.3      & 28.5      & 36.2        & 25.9         & 20.6         & 7.2  & 8.9  & 11.6      & 16.8 & 28.6 \\
MI-FGSM      & 100.0    & 66.6      & 71.1      & 77.7        & 54.8         & 50.6         & 18.6 & 25.5 & 35.3      & 42.7 & 54.3 \\
Admix        & 100.0    & 89.6      & 90.5      & 94.6        & 80.3         & 77.3         & 31.8 & 38.5 & 56.0      & 60.4 & 71.9 \\
BSR & 100.0    & 95.8      & 96.6      & 98.1        & 88.9         & 90.2         & 46.1 & 58.7 & 77.7      & 77.6 & 83.0 \\
DEM          & 100.0    & 95.5      & 95.8      & 98.1        & 92.2         & 90.4         & 46.9 & 45.0 & 67.7      & 64.3 & 79.6 \\
DIM          & 100.0    & 84.6      & 87.8      & 93.6        & 77.6         & 73.3         & 31.1 & 37.7 & 53.1      & 56.8 & 69.6 \\
SIA         & 100.0    & 96.5      & 97.1      & 98.6        & 90.0         & 89.2         & 44.4 & 56.8 & 74.3      & 76.0 & 82.3 \\
IDE          & 99.9     & 66.0      & 68.4      & 75.5        & 56.3         & 51.3         & 18.8 & 23.4 & 34.2      & 40.9 & 53.5 \\
Masked       & 100.0    & 71.6      & 76.2      & 80.5        & 58.7         & 54.7         & 20.1 & 26.1 & 37.4      & 44.4 & 57.0 \\
SIM          & 100.0    & 83.0      & 85.9      & 90.7        & 74.0         & 69.3         & 26.2 & 35.2 & 48.4      & 52.4 & 66.5 \\
$\rm{S}^2$IM          & 100.0    & 90.4      & 92.6      & 94.1        & 83.8         & 80.4         & 32.9 & 41.6 & 56.2      & 62.4 & 73.4 \\
TIM          & 100.0    & 58.7      & 67.4      & 72.4        & 52.1         & 48.6         & 18.3 & 17.4 & 26.8      & 34.6 & 49.6 \\
ATTA         & 88.0     & 47.9      & 50.1      & 58.3        & 42.7         & 35.4         & 14.0 & 17.7 & 24.6      & 30.7 & 40.9 \\
AutoMA       & 100      & 93.2      & 95.1      & 97.4        & 86.4         & 87.0           & 41   & 50.7 & 67.7      & 67.8 & 78.6 \\
AITL         & 99.6     & 93.3      & 95.2      & 96.8        & 91.8         & 91.2         & 47.5 & 51.8 & 68.9      & 71.2 & 80.7 \\
\rowcolor{mygray} L2T (Ours)           & 100.0      & 99.3      & 99.2      & 99.6        & 96.9         & 97.4         & 63.7 & 71.1 & 86.6      & 86.0   & 90.0   \\
\bottomrule
\end{tabular}}
\label{tabs:resnet18}
\end{table*}
\begin{table*}[hbp]
\setlength\tabcolsep{3pt}
\caption{Attack success rate (\%) across ten models on the adversarial examples crafted on ResNet-101 by different attack}
\resizebox{\linewidth}{!}{
\begin{tabular}{cccccccccccc}
\toprule
Attack       & Res-18 & Res-101 & NeXt-50 & Dense-121 & Inc-v3 & Inc-v4 & ViT  & PiT  & Visformer & Swin & Average \\
\midrule
I-FGSM       & 36.6   & 100.0   & 35.4    & 33.2      & 25.8   & 20.6   & ~8.0  & 10.3 & 13.0      & 16.3 & 29.9    \\
MI-FGSM      & 72.6   & 100.0   & 73.8    & 71.7      & 54.1   & 49.6   & 22.7 & 27.2 & 34.5      & 38.3 & 54.4    \\
Admix        & 94.6   & 100.0   & 94.0    & 94.6      & 82.9   & 78.0   & 38.2 & 46.9 & 57.9      & 60.3 & 74.7    \\
BSR & 97.4   & 100.0   & 97.9    & 97.8      & 89.2   & 90.9   & 56.4 & 67.4 & 80.6      & 81.1 & 85.9    \\
DEM         & 97.6   & 100.0   & 96.8    & 97.5      & 91.7   & 89.5   & 52.2 & 51.9 & 66.8      & 68.4 & 81.2    \\
DIM          & 86.0   & 99.9    & 89.9    & 89.3      & 75.1   & 74.5   & 38.5 & 45.6 & 56.8      & 57.3 & 71.3    \\
SIA         & 98.1   & 100.0   & 97.9    & 98.0      & 87.8   & 89.4   & 48.9 & 58.9 & 75.0      & 74.3 & 82.8    \\
IDE          & 78.5   & 96.4    & 72.8    & 73.6      & 59.9   & 56.6   & 23.8 & 25.6 & 34.7      & 43.0 & 56.5    \\
Masked       & 80.9   & 100.0   & 80.9    & 80.2      & 58.8   & 54.5   & 25.0 & 30.4 & 40.2      & 43.2 & 59.4    \\
SIM          & 86.8   & 100.0   & 88.0    & 89.2      & 74.9   & 68.7   & 33.1 & 39.1 & 50.1      & 51.7 & 68.2    \\
$\rm{S}^2$IM            & 95.9   & 100.0   & 94.8    & 94.7      & 88.3   & 84.3   & 45.7 & 51.7 & 62.3      & 67.1 & 78.5    \\
TIM          & 69.3   & 100.0   & 72.8    & 67.2      & 50.9   & 47.8   & 23.2 & 23.2 & 30.7      & 36.8 & 52.2    \\
ATTA         & 51.7   & 73.1    & 50.7    & 49.6      & 41.2   & 35.8   & 15.9 & 19.8 & 25.4      & 27.8 & 39.9    \\
AutoMA         & 95.5   & 99.7    & 95.4    & 95.2      & 85.6   & 86.1   & 50.5 & 59.8 & 70.3      & 70.9 & 80.9    \\
AITL         & 96.6   & 99.1    & 96.5    & 97.8      & 92.0     & 92.5   & 57.1 & 64.9 & 76.0        & 76.3 & 84.9    \\
\rowcolor{mygray} L2T (Ours)           & 99.3   & 100.0     & 99.2    & 99.5      & 97.1   & 96.8   & 72.3 & 77.9 & 88.9      & 88.1 & 91.9   \\
\bottomrule
\end{tabular}}
\label{tabs:resnet101}
\end{table*}
\begin{table*}
\setlength\tabcolsep{3pt}
\caption{Attack success rate (\%) across ten models on the adversarial examples crafted on DenseNet-121 by different attack}
\resizebox{\linewidth}{!}{
\begin{tabular}{cccccccccccc}
\toprule
Attack       & Res-18 & Res-101 & NeXt-50 & Dense-121 & Inc-v3 & Inc-v4 & ViT  & PiT  & Visformer & Swin & Average \\ \midrule
I-FGSM       & 44.5   & 34.0    & 36.6    & 100.0     & 28.6   & 23.9   & 8.1  & 11.3 & 14.7      & 20.8 & 32.2    \\
MI-FGSM      & 78.6   & 68.9    & 74.8    & 100.0     & 56.6   & 53.6   & 24.5 & 31.1 & 44.0      & 45.6 & 57.8    \\
Admix        & 94.3   & 91.1    & 93.4    & 100.0     & 82.5   & 81.1   & 40.8 & 50.7 & 68.3      & 65.8 & 76.8    \\
BSR & 97.4   & 85.7    & 97.3    & 100.0     & 89.7   & 91.5   & 52.2 & 68.3 & 84.7      & 80.0 & 84.7    \\
DEM         & 97.8   & 94.5    & 97.1    & 100.0     & 92.2   & 91.5   & 53.8 & 56.0 & 74.4      & 70.8 & 82.8    \\
DIM          & 88.4   & 84.1    & 89.7    & 100.0     & 76.4   & 75.5   & 36.5 & 44.0 & 62.0      & 59.5 & 71.6    \\
SIA & 98.4   & 96.4    & 97.5    & 100.0     & 89.1   & 92.8   & 49.7 & 64.1 & 83.4      & 78.1 & 85.0    \\
IDE          & 87.8   & 77.3    & 80.6    & 99.4      & 70.6   & 68.5   & 26.3 & 35.0 & 49.5      & 51.8 & 64.7    \\
Masked       & 82.8   & 74.0    & 81.2    & 100.0     & 60.6   & 60.8   & 25.7 & 35.7 & 49.3      & 51.3 & 62.1    \\
SIM          & 89.7   & 84.2    & 88.3    & 100.0     & 75.3   & 74.2   & 32.6 & 42.8 & 59.2      & 57.3 & 70.4    \\
$\rm{S}^2$IM            & 97.2   & 94.9    & 96.9    & 100.0     & 90.7   & 90.2   & 50.7 & 61.6 & 78.5      & 76.9 & 83.8    \\
TIM          & 74.7   & 62.4    & 70.9    & 100.0     & 52.2   & 51.6   & 20.1 & 21.7 & 33.9      & 38.9 & 52.6    \\
ATTA         & 54.8   & 45.6    & 49.7    & 79.4      & 42.2   & 36.8   & 15.3 & 20.6 & 28.3      & 32.3 & 40.5    \\
AutoMA         & 95.3   & 93.8    & 95.2    & 99.9      & 85.4   & 86.9   & 46.5 & 59.6 & 73.0        & 71.3 & 80.7    \\
AITL         & 97.1   & 94.3    & 96.0      & 99.5      & 91.3   & 92.6   & 53.7 & 61.5 & 76.0        & 74.6 & 83.7    \\
\rowcolor{mygray} L2T (Ours)           & 99.5   & 98.9    & 99.3    & 100.0       & 97.4   & 98.3   & 71.3 & 79.7 & 92.9      & 90.2 & 92.8   \\ \bottomrule
\end{tabular}}
\label{tabs:dense121}
\end{table*}
\begin{table*}[hbp]
\setlength\tabcolsep{3pt}
\caption{Attack success rate (\%) across ten models on the adversarial examples crafted on ResNeXt-50 by different attack}
\resizebox{\linewidth}{!}{
\begin{tabular}{cccccccccccc} 
\\ \toprule
Attack  & Res-18 & Res-101 & NeXt-50 & Dense-121 & Inc-v3 & Inc-v4 & ViT  & PiT  & Visformer & Swin & Average \\ \midrule
I-FGSM  & 32.4   & 29.4    & 99.4    & 31.8      & 25.0   & 18.5   & ~7.3  & ~9.8  & 13.1      & 15.8 & 28.2    \\
MI-FGSM & 64.7   & 62.9    & 99.9    & 69.2      & 49.3   & 45.7   & 19.1 & 27.0 & 35.6      & 38.8 & 51.2    \\
Admix   & 88.7   & 87.4    & 100.0   & 94.3      & 78.0   & 73.7   & 33.6 & 44.0 & 58.5      & 57.3 & 71.5    \\
BSR     & 95.8   & 95.7    & 100.0   & 97.5      & 83.3   & 86.9   & 47.9 & 66.8 & 79.5      & 74.5 & 82.8    \\
DEM     & 96.6   & 94.8    & 100.0   & 97.9      & 89.5   & 90.5   & 49.5 & 55.1 & 70.9      & 67.5 & 81.2    \\
DIM     & 81.7   & 80.7    & 99.8    & 85.1      & 67.7   & 69.0   & 33.7 & 42.4 & 53.1      & 54.2 & 66.7    \\
SIA     & 97.0   & 95.1    & 100.0   & 97.2      & 83.5   & 85.8   & 44.6 & 60.6 & 76.9      & 73.7 & 81.4    \\
IDE     & 76.2   & 66.1    & 96.3    & 71.0      & 54.8   & 55.0   & 20.7 & 26.8 & 36.1      & 42.6 & 54.6    \\
Masked  & 74.8   & 70.6    & 100.0   & 76.1      & 52.5   & 50.8   & 22.3 & 31.2 & 41.2      & 43.3 & 56.3    \\
SIM     & 79.3   & 76.9    & 100.0   & 86.3      & 66.2   & 62.2   & 25.9 & 36.6 & 48.0      & 47.5 & 62.9    \\
$\rm{S}^2$IM      & 95.5   & 94.3    & 99.9    & 96.6      & 86.2   & 85.3   & 45.5 & 56.3 & 67.3      & 71.4 & 79.8    \\
TIM     & 65.6   & 58.6    & 99.8    & 64.3      & 45.5   & 44.2   & 18.4 & 20.9 & 30.1      & 37.7 & 48.5    \\
ATTA    & 43.1   & 39.8    & 66.9    & 42.9      & 34.3   & 29.9   & 14.0 & 17.5 & 22.9      & 25.1 & 33.6    \\
AutoMA  & 89.6   & 91.0      & 99.7    & 93.4      & 78.4   & 80.8   & 42.3 & 57.7 & 67.7      & 66.9 & 76.8    \\
AITL    & 94.0     & 92.4    & 98.9    & 96.6      & 88.7   & 88.9   & 47.5 & 59.8 & 72.5      & 70.1 & 80.9    \\
\rowcolor{mygray} L2T (Ours)     & 99.4   & 99.2    & 100.0     & 99.3      & 95.6   & 97.2   & 67.2 & 78.2 & 88.1      & 85.8 & 91.0     \\ \bottomrule
\end{tabular}}
\label{tabs:next50}
\end{table*}
\begin{table*}[]
\setlength\tabcolsep{3pt}
\caption{Attack success rate (\%) across ten models on the adversarial examples crafted on Inception-v3 by different attack}
\resizebox{\linewidth}{!}{
\begin{tabular}{cccccccccccc}
\toprule
Attack  & Res-18 & Res-101 & NeXt-50 & Dense-121 & Inc-v3 & Inc-v4 & ViT  & PiT  & Visformer & Swin & Average \\ \midrule
I-FGSM  & 19.7   & 13.7    & 14.6    & 16.8      & 98.5   & 21.9   & ~6.7  & ~7.7  & ~8.8       & 13.4 & 22.2    \\
MI-FGSM & 48.0   & 37.5    & 38.5    & 42.9      & 98.7   & 49.3   & 16.4 & 20.7 & 23.8      & 29.0 & 40.5    \\
Admix   & 66.7   & 57.6    & 58.5    & 67.2      & 99.8   & 76.5   & 23.5 & 28.8 & 34.4      & 41.1 & 55.4    \\
BSR     & 88.4   & 81.9    & 84.3    & 88.2      & 99.8   & 91.7   & 39.3 & 48.4 & 60.8      & 64.0 & 74.7    \\
DEM     & 77.5   & 68.7    & 71.4    & 75.3      & 99.5   & 85.0   & 34.8 & 34.1 & 43.7      & 50.5 & 64.0    \\
DIM     & 59.4   & 48.2    & 51.7    & 57.4      & 99.0   & 66.4   & 21.5 & 24.3 & 31.2      & 37.9 & 49.7    \\
SIA     & 82.9   & 73.0    & 76.0    & 81.6      & 99.3   & 88.2   & 31.9 & 41.4 & 51.7      & 55.6 & 68.2    \\
IDE     & 56.4   & 41.9    & 44.9    & 46.5      & 95.4   & 56.7   & 15.6 & 19.1 & 23.0      & 29.3 & 42.9    \\
Masked  & 55.7   & 45.8    & 45.1    & 50.4      & 100.0  & 58.3   & 17.5 & 22.7 & 27.3      & 32.8 & 45.6    \\
SIM     & 60.2   & 47.7    & 46.8    & 54.1      & 99.8   & 64.2   & 19.6 & 23.7 & 26.4      & 33.1 & 47.6    \\
$\rm{S}^2$IM       & 71.5   & 64.5    & 66.1    & 70.7      & 99.6   & 82.7   & 27.6 & 36.4 & 42.1      & 50.2 & 61.1    \\
TIM     & 44.6   & 31.7    & 37.6    & 38.9      & 98.2   & 42.3   & 13.5 & 13.3 & 16.2      & 23.0 & 35.9    \\
ATTA    & 31.0     & 21.0      & 22.1    & 23.8      & 50.9   & 28     & 10.4 & 11.6 & 13.3      & 19.2 & 23.1    \\
AutoMA  & 65.6   & 58.0      & 62.2    & 65.6      & 98.5   & 76.1   & 27.1 & 32.6 & 38.8      & 44.2 & 56.7    \\
AITL    & 77.1   & 69.9    & 72.2    & 79.6      & 98.9   & 85.8   & 34.3 & 38.9 & 46.6      & 53.4 & 65.7    \\
\rowcolor{mygray} L2T (Ours)      & 89.9   & 86.5    & 88.1    & 91.9      & 99.6   & 94.8   & 48.7 & 54.1 & 65.4      & 69.3 & 78.8   \\ \bottomrule
\end{tabular}}
\label{tabs:inc-v3}
\end{table*}
\begin{table*}[]
\setlength\tabcolsep{3pt}
\caption{Attack success rate (\%) across ten models on the adversarial examples crafted on Inception-v4 by different attack}
\resizebox{\linewidth}{!}{
\begin{tabular}{cccccccccccc}
\toprule
Attack  & Res-18 & Res-101 & NeXt-50 & Dense-121 & Inc-v3 & Inc-v4 & ViT  & PiT  & Visformer & Swin & Average \\ \midrule
I-FGSM  & 22.4   & 15.0    & 17.3    & 18.4      & 30.5   & 95.7   & ~6.3  & ~8.6  & 11.4      & 13.9 & 23.9    \\
MI-FGSM & 50.1   & 41.3    & 43.7    & 47.6      & 58.2   & 97.1   & 17.4 & 21.4 & 28.4      & 31.5 & 43.7    \\
Admix   & 74.9   & 69.0    & 71.7    & 78.6      & 88.2   & 99.7   & 33.3 & 39.4 & 50.6      & 52.8 & 65.8    \\
BSR     & 87.3   & 79.1    & 85.6    & 89.3      & 89.3   & 99.9   & 38.5 & 52.4 & 66.6      & 65.2 & 75.3    \\
DEM     & 79.0   & 71.0    & 76.2    & 79.4      & 87.9   & 99.2   & 35.6 & 37.4 & 52.3      & 52.8 & 67.1    \\
DIM     & 63.0   & 55.4    & 60.4    & 63.8      & 73.2   & 96.8   & 24.7 & 31.5 & 39.6      & 40.8 & 54.9    \\
SIA     & 83.0   & 73.3    & 78.5    & 85.5      & 87.6   & 99.7   & 34.1 & 44.6 & 59.0      & 59.8 & 70.5    \\
IDE     & 56.8   & 45.8    & 48.5    & 54.9      & 64.2   & 92.5   & 17.4 & 23.3 & 28.0      & 33.6 & 46.5    \\
Masked  & 56.0   & 47.7    & 49.3    & 57.3      & 65.2   & 99.7   & 19.9 & 26.1 & 33.9      & 36.5 & 49.2    \\
SIM     & 66.3   & 60.2    & 64.4    & 71.1      & 80.8   & 99.5   & 28.9 & 35.0 & 44.0      & 44.6 & 59.5    \\
$\rm{S}^2$IM      & 76.5   & 69.9    & 72.9    & 77.8      & 85.4   & 99.4   & 33.6 & 42.4 & 50.6      & 54.7 & 66.3    \\
TIM     & 46.6   & 35.8    & 41.6    & 44.1      & 50.8   & 96.2   & 13.3 & 14.8 & 19.0      & 24.5 & 38.7    \\
ATTA    & 32.6   & 24.1    & 25.6    & 28.4      & 36.2   & 46.2   & 11.3 & 13.3 & 17.0        & 20.0   & 25.5    \\
AutoMA  & 71.8   & 63.8    & 69.4    & 75.1      & 84.1   & 97.9   & 32   & 39.5 & 50.3      & 49.8 & 63.4    \\
AITL    & 81.1   & 75.3    & 79.4    & 86.1      & 90.8   & 99.3   & 41   & 47.3 & 59.5      & 59.2 & 71.9    \\
\rowcolor{mygray}  L2T (Ours)      & 91.5   & 88.8    & 91.1    & 94.5      & 95.4   & 99.9   & 51.7 & 61.9 & 75.1      & 74.0   & 82.4   \\ \bottomrule
\end{tabular}}
\label{tabs:Inc-v4}
\end{table*}
\begin{table*}[]
\setlength\tabcolsep{3pt}
\caption{Attack success rate (\%) across ten models on the adversarial examples crafted on ViT by different attacks}
\resizebox{\linewidth}{!}{
\begin{tabular}{cccccccccccc}
\toprule
Attack  & Res-18 & Res-101 & NeXt-50 & Dense-121 & Inc-v3 & Inc-v4 & ViT  & PiT  & Visformer & Swin & Average \\ \midrule
I-FGSM  & 26.3   & 19.8    & 21.7    & 23.6      & 23.4   & 20.6   & 99.7 & 20.0 & 20.6      & 33.1 & 30.9    \\
MI-FGSM & 52.9   & 44.7    & 48.3    & 51.3      & 45.6   & 42.2   & 99.7 & 44.6 & 45.7      & 60.6 & 53.6    \\
Admix   & 64.9   & 59.8    & 61.2    & 64.1      & 62.1   & 57.3   & 99.2 & 60.6 & 62.2      & 74.4 & 66.6    \\
BSR     & 83.6   & 83.8    & 86.2    & 87.8      & 79.9   & 81.8   & 99.7 & 90.3 & 90.4      & 89.6 & 87.3    \\
DEM     & 76.6   & 78.5    & 80.8    & 81.8      & 79.6   & 79.0   & 99.9 & 82.1 & 81.7      & 81.0 & 82.1    \\
DIM     & 63.2   & 60.7    & 62.5    & 65.3      & 61.1   & 59.8   & 98.7 & 66.5 & 64.1      & 71.4 & 67.3    \\
SIA     & 82.0   & 79.9    & 82.0    & 83.4      & 75.2   & 78.1   & 99.7 & 85.4 & 85.8      & 88.4 & 84.0    \\
IDE     & 67.1   & 60.8    & 64.2    & 66.3      & 62.5   & 59.7   & 99.3 & 56.8 & 58.8      & 72.6 & 66.8    \\
Masked  & 55.6   & 47.5    & 50.9    & 54.8      & 49.3   & 44.5   & 99.8 & 49.2 & 49.7      & 65.6 & 56.7    \\
SIM     & 60.8   & 53.0    & 55.6    & 60.8      & 55.1   & 51.7   & 99.3 & 53.7 & 56.4      & 68.4 & 61.5    \\
$\rm{S}^2$IM       & 67.8   & 63.2    & 65.6    & 69.4      & 68.3   & 65.5   & 99.9 & 66.7 & 67.3      & 78.3 & 71.2    \\
TIM     & 49.1   & 42.3    & 46.3    & 47.1      & 40.3   & 37.6   & 98.9 & 34.5 & 37.7      & 46.5 & 48.0    \\
ATTA    & 41.9   & 33.6    & 36.1    & 39.3      & 39.3   & 32.9   & 79.8 & 32.7 & 32.6      & 42.0   & 41.0      \\
AutoMA  & 72.1   & 71.0      & 73.0      & 75.8      & 70.9   & 71.4   & 97.9 & 77.9 & 77.6      & 78.6 & 76.6    \\
AITL    & 76.8   & 74.4    & 77.7    & 78.6      & 77.7   & 75.8   & 94.9 & 79.5 & 78.9      & 79.6 & 79.4    \\
\rowcolor{mygray} L2T (Ours)      & 89.7   & 87.3    & 88.7    & 89.6      & 87.4   & 86.8   & 98.2 & 90.6 & 90.8      & 92.3 & 90.1   \\ \bottomrule
\end{tabular}}
\label{tabs:vit}
\end{table*}
\begin{table*}[]
\setlength\tabcolsep{3pt}
\caption{Attack success rate (\%) across ten models on the adversarial examples crafted on PiT by different attacks}
\resizebox{\linewidth}{!}{
\begin{tabular}{cccccccccccc}
\toprule
Attack  & Res-18 & Res-101 & NeXt-50 & Dense-121 & Inc-v3 & Inc-v4 & ViT  & PiT  & Visformer & Swin & Average \\ \midrule
I-FGSM  & 22.1   & 15.9    & 18.4    & 19.9      & 23.3   & 17.7   & 11.3 & 85.1 & 21.6      & 24.8 & 26.0    \\
MI-FGSM & 52.3   & 41.8    & 48.3    & 51.8      & 46.4   & 43.0   & 30.9 & 97.6 & 53.1      & 55.9 & 52.1    \\
Admix   & 63.0   & 55.1    & 61.8    & 63.5      & 57.3   & 56.8   & 46.7 & 97.5 & 67.5      & 70.4 & 64.0    \\
BSR     & 80.9   & 77.6    & 84.0    & 85.0      & 74.7   & 76.8   & 70.9 & 99.2 & 89.5      & 90.0 & 82.9    \\
DEM     & 79.4   & 74.7    & 78.5    & 80.5      & 78.3   & 76.9   & 68.7 & 99.9 & 84.9      & 83.0 & 80.5    \\
DIM     & 63.3   & 58.7    & 64.6    & 64.8      & 61.5   & 62.4   & 50.9 & 94.3 & 70.1      & 71.7 & 66.2    \\
SIA     & 81.3   & 77.2    & 85.6    & 84.9      & 75.8   & 77.3   & 69.7 & 99.0 & 90.6      & 91.6 & 83.3    \\
IDE     & 68.8   & 61.5    & 64.0    & 68.4      & 66.1   & 64.0   & 53.1 & 94.2 & 70.2      & 71.2 & 68.2    \\
Masked  & 59.1   & 51.7    & 57.2    & 59.0      & 53.5   & 49.1   & 39.1 & 99.3 & 61.8      & 63.9 & 59.4    \\
SIM     & 62.0   & 54.2    & 59.9    & 61.6      & 55.7   & 53.6   & 43.6 & 99.2 & 65.1      & 68.5 & 62.3    \\
$\rm{S}^2$IM     & 71.6   & 68.9    & 70.9    & 73.8      & 71.7   & 69.9   & 61.2 & 96.4 & 76.1      & 78.3 & 73.9    \\
TIM     & 48.7   & 37.9    & 47.7    & 47.3      & 40.7   & 37.7   & 27.9 & 93.8 & 42.2      & 48.0 & 47.2    \\
ATTA    & 44.4   & 32.1    & 38.1    & 40.3      & 39.7   & 35.4   & 23.7 & 71.6 & 37.6      & 40.2 & 40.3    \\
AutoMA  & 71.1   & 67.9    & 74.8    & 76.2      & 69.8   & 67.5   & 62.8 & 96.6 & 80.4      & 81.2 & 74.8    \\
AITL    & 79.6   & 79.0      & 82.5    & 83.5      & 81.2   & 80.1   & 74.6 & 93.5 & 86.7      & 86.4 & 82.7    \\
\rowcolor{mygray} L2T (Ours)      & 93.2   & 90.1    & 93.0      & 94.3      & 90.7   & 90.7   & 89.8 & 99.5 & 96.9      & 97.1 & 93.5   \\ \bottomrule
\end{tabular}}
\label{tabs:pit}
\end{table*}
\begin{table*}[]
\setlength\tabcolsep{3pt}
\caption{Attack success rate (\%) across ten models on the adversarial examples crafted on Visformer by different attacks}
\resizebox{\linewidth}{!}{
\begin{tabular}{cccccccccccc}
\toprule
Attack  & Res-18 & Res-101 & NeXt-50 & Dense-121 & Inc-v3 & Inc-v4 & ViT  & PiT  & Visformer & Swin & Average \\ \midrule
I-FGSM  & 25.4   & 20.9    & 24.4    & 26.6      & 25.4   & 21.4   & 12.0 & 22.4 & 93.3      & 32.6 & 30.2    \\
MI-FGSM & 59.8   & 50.1    & 55.3    & 60.2      & 50.2   & 50.8   & 34.5 & 54.6 & 98.3      & 64.3 & 57.8    \\
Admix   & 77.1   & 70.0    & 77.4    & 80.0      & 69.4   & 71.0   & 55.4 & 77.3 & 97.8      & 83.7 & 75.9    \\
BSR     & 86.0   & 82.9    & 88.8    & 90.5      & 79.5   & 83.7   & 65.7 & 90.4 & 99.5      & 91.7 & 85.9    \\
DEM     & 84.3   & 81.4    & 86.6    & 87.8      & 83.5   & 85.1   & 65.8 & 83.0 & 99.9      & 85.0 & 84.3    \\
DIM     & 71.9   & 68.5    & 74.9    & 79.1      & 69.2   & 70.5   & 52.2 & 75.1 & 96.8      & 79.5 & 73.8    \\
SIA     & 86.6   & 84.5    & 89.9    & 91.7      & 80.2   & 84.2   & 69.7 & 90.9 & 98.9      & 92.8 & 86.9    \\
IDE     & 77.9   & 71.6    & 75.8    & 79.6      & 73.5   & 73.8   & 57.4 & 73.7 & 97.0      & 81.2 & 76.2    \\
Masked  & 63.5   & 54.3    & 61.4    & 64.6      & 54.7   & 54.6   & 37.1 & 60.0 & 99.2      & 68.5 & 61.8    \\
SIM     & 71.1   & 65.7    & 71.2    & 75.3      & 64.5   & 66.5   & 49.5 & 71.6 & 97.8      & 79.6 & 71.3    \\
$\rm{S}^2$IM       & 82.1   & 78.3    & 81.6    & 86.1      & 81.6   & 82.2   & 66.4 & 81.7 & 97.2      & 87.3 & 82.5    \\
TIM     & 57.4   & 47.7    & 56.9    & 58.9      & 46.6   & 47.5   & 33.9 & 48.1 & 97.6      & 60.0 & 55.5    \\
ATTA    & 50.0     & 39.5    & 45.7    & 49.5      & 41.5   & 41.8   & 26.8 & 42.8 & 85.9      & 51.8 & 47.5    \\
AutoMA  & 79.3   & 78.0      & 85.4    & 86.7      & 77.3   & 80.9   & 66.8 & 85.4 & 98.2      & 87.8 & 82.6    \\
AITL    & 87.2   & 85.0      & 88.4    & 89.3      & 84.1   & 87.0     & 76.6 & 88.7 & 96.5      & 90.5 & 87.3    \\
\rowcolor{mygray} L2T (Ours)      & 96.8   & 95.6    & 97.1    & 97.9      & 94.4   & 96.5   & 89.9 & 96.6 & 100.0       & 97.5 & 96.2   \\ \bottomrule
\end{tabular}}
\label{tabs:vis}
\end{table*}
\begin{table*}[]
\setlength\tabcolsep{3pt}
\caption{Attack success rate (\%) across ten models on the adversarial examples crafted on Swin by different attacks}
\resizebox{\linewidth}{!}{
\begin{tabular}{cccccccccccc}
\toprule
Attack  & Res-18 & Res-101 & NeXt-50 & Dense-121 & Inc-v3 & Inc-v4 & ViT  & PiT  & Visformer & Swin  & Average \\ \midrule
I-FGSM  & 14.3   & 10.8    & ~9.9     & 13.2      & 17.5   & 11.6   & ~5.9  & ~8.1  & 10.8      & 72.3  & 17.4    \\
MI-FGSM & 44.9   & 32.6    & 36.6    & 39.9      & 37.1   & 31.7   & 22.5 & 32.0 & 40.1      & 98.8  & 41.6    \\
Admix   & 56.0   & 41.6    & 47.2    & 51.7      & 45.0   & 41.6   & 31.4 & 43.8 & 53.7      & 99.2  & 51.1    \\
BSR     & 86.9   & 79.1    & 86.3    & 87.3      & 76.4   & 78.6   & 65.6 & 88.8 & 92.0      & 99.3  & 84.0    \\
DEM     & 79.4   & 75.6    & 78.3    & 80.0      & 76.5   & 77.2   & 61.5 & 79.1 & 81.4      & 100.0 & 78.9    \\
DIM     & 70.9   & 64.8    & 70.4    & 72.0      & 66.8   & 67.3   & 52.3 & 73.4 & 76.4      & 98.0  & 71.2    \\
SIA     & 82.7   & 74.5    & 79.3    & 84.2      & 70.5   & 72.1   & 59.3 & 82.5 & 88.7      & 99.1  & 79.3    \\
IDE     & 67.3   & 54.8    & 59.1    & 63.9      & 61.4   & 56.8   & 43.8 & 54.2 & 61.9      & 98.4  & 62.2    \\
Masked  & 46.5   & 33.4    & 39.7    & 43.8      & 39.7   & 33.2   & 26.7 & 35.0 & 44.8      & 99.5  & 44.2    \\
SIM     & 53.0   & 38.3    & 44.6    & 48.2      & 42.2   & 40.4   & 29.9 & 39.9 & 49.5      & 99.2  & 48.5    \\
$\rm{S}^2$IM     & 83.4   & 75.6    & 80.1    & 83.9      & 77.9   & 79.2   & 67.8 & 80.8 & 85.7      & 99.1  & 81.4    \\
TIM     & 58.7   & 46.9    & 58.0    & 58.9      & 48.1   & 46.2   & 33.5 & 45.0 & 51.7      & 99.0  & 54.6    \\
ATTA    & 38.3   & 28.1    & 32.1    & 34.6      & 34.6   & 28.2   & 20.3 & 28.2 & 34.9      & 92.0    & 37.1    \\
AutoMA  & 81.9   & 78.2    & 83.3    & 84.5      & 76.0     & 78.0     & 65.7 & 86.9 & 89.0        & 98.7  & 82.2    \\
AITL    & 87.8   & 84.0      & 89.8    & 90.9      & 86.9   & 88.5   & 72.0   & 89.4 & 90.5      & 97.1  & 87.7    \\
\rowcolor{mygray} L2T (Ours)      & 94.4   & 91.9    & 94.2    & 95.9      & 90.7   & 93.1   & 85.9 & 94.5 & 96.3      & 99.6  & 93.6    \\ \bottomrule
\end{tabular}}
\label{tabs:swin}
\end{table*}

%\subsection{Detailed Results of \cref{fig:defense}}
\begin{table*}[hbp]
\centering
\caption{Attack success rate(\%) on adversarial examples on ensemble attack across four defense methods and four vision API.}
\begin{tabular}{ccccccccc}
\toprule
Attack & AT   & HGD  & NRP  & RS   & Google & Azure & GPT-4V & Bard \\ \midrule
SIM   & 36.3 & 83.8 & 65.7 & 26.4 & 77.5 & 69.8  & 62.4 & 79.7 \\
TIM   & 36.6 & 63.8 & 56.0   & 35.7 & 55.3 & 52.6  & 64.1 & 71.4 \\
Admix & 37.8 & 91.1 & 70.8 & 29.4 & 73.6 & 57.1  & 76.0 & 83.2 \\
DEM   & 40.3 & 88.9 & 74.9 & 37.8 & 76.4 & 69.3  & 83.3 & 91.3 \\
AutoMA & 37.9 & 89.1 & 66.5 & 30.0   & 67.4 & 61.9  & 71.4 & 86.2 \\
IDE   & 40.9 & 73.1 & 68.0   & 38.0   & 71.0 & 64.8  & 57.1 & 73.1 \\
ATTA  & 30.3 & 49.9 & 47.8 & 18.4 & 49.0 & 47.9  & 39.4 & 75.9 \\
Masked & 32.6 & 72.9 & 49.6 & 21.1 & 57.3 & 52.7  & 72.0 & 84.3 \\
AITL  & 44.3 & 91.1 & 79.9 & 42.1 & 79.4 & 65.2  & 79.6 & 90.2 \\
$\rm{S}^2$IM   & 41.1 & 90.6 & 80.1 & 37.0   & 67.0 & 65.1  & 86.2 & 93.6 \\
BSR   & 38.7 & 92.6 & 63.4 & 29.7 & 74.4 & 55.8  & 82.5 & 95.1 \\
SIA  & 37.6 & 91.5 & 63.1 & 28.9 & 77.5 & 69.1  & 89.6 & 94.2 \\
\rowcolor{mygray} L2T (Ours)  & 47.9 & 98.5 & 87.2 & 46.7 & 86.5 & 82.7  & 96.7 & 99.9 \\ \bottomrule
\end{tabular}
\label{tabs:defense&API}
\end{table*}

%\subsection{Detailed Results of \cref{fig:steps}}
\begin{table*}[h!]
\centering
\caption{Attack success rate(\%) on adversarial examples crafts on ResNet-18 by different iterations.}
\resizebox{\textwidth}{!}{
\begin{tabular}{cccccccccccccc}
\toprule
Iteration & SIM   & TIM   & Admix & DEM   & AutoMA & IDE   & ATTA  & Masked  & AITL  & $\rm{S}^2$IM & BSR & SIA   & L2T(Ours)   \\ \midrule
1         & ~9.1   & 12.5 & ~7.9  & 60.3 & ~8.5   & ~7.3  & ~7.7  & ~9.3  &~ 7.7  & ~6.6  & ~8.5      & ~7.4  & ~8.4  \\
2         & 19.7 & 20.2 & 19.2 & 71.6 & 22.9  & 13.1 & 13.2 & 20.8  & 18.7 & 13.6 & 25.5     & 19.3 & 23.5 \\
3         & 25.2 & 24.4 & 26.2& 74.2 & 31.5  & 17.1 & 16.0 & 24.8 & 26.9 & 19.9 & 35.4     & 28.7 & 34.1 \\
4         & 35.9 & 29.8 & 38.1 & 76.0 & 45.5  & 24.0 & 21.3 & 33.0  & 41.8 & 33.2 & 51.1     & 44.0 & 51.3 \\
5         & 42.0 & 33.5 & 45.4 & 76.3 & 53.4  & 29.1 & 24.8 & 37.9 & 50.6 & 41.4 & 59.7     & 52.9 & 60.9 \\
6         & 48.8 & 37.7 & 53.3 & 77.6 & 61.0  & 35.3 & 28.6  & 43.0 & 59.0 & 50.8 & 68.1     & 62.4 & 70.8 \\
7         & 55.5 & 41.9 & 60.4 & 77.7 & 67.7  & 41.0 & 32.5 & 48.0 & 66.8 & 59.7 & 74.5     & 70.2 & 79.1 \\
8         & 58.3 & 44.1 & 64.2 & 78.3 & 71.7  & 44.4 & 35.3 & 50.3 & 71.8 & 63.8 & 77.3     & 74.1 & 83.1 \\
9      & 63.1 & 47.3 & 68.9 & 79.0 & 75.9  & 50.2 & 38.7 & 54.7 & 77.8 & 70.1 & 81.9     & 79.4 & 87.3 \\
10     & 66.1 & 49.3 & 71.5 & 79.0 & 78.6  & 53.7 & 40.9 & 57.0 & 81.0 & 73.4 & 83.9     & 82.9 & 89.4 \\
20     & 67.2 & 50.1 & 72.0 & 81.3 & 78.8  & 57.9 & 44.7 & 57.2 & 81.3 & 72.6 & 83.0     & 84.3 & 91.4 \\
30     & 67.0 & 50.9 & 71.6 & 82.2 & 79.1  & 57.6 & 44.6 & 56.4 & 81.5 & 71.2 & 82.2     & 83.7 & 91.5 \\
40     & 67.4 & 51.2 & 71.6 & 82.8 & 79.4  & 58.6 & 45.1 & 55.8 & 81.4 & 71.4 & 83.0     & 84.1 & 91.8 \\
50     & 67.5 & 51.6 & 71.9 & 82.7 & 80.1  & 59.2 & 45.3 & 56.2 & 83.2 & 70.7 & 83.5     & 84.4 & 92.3 \\
60     & 67.4 & 51.9 & 71.6 & 83.0 & 80.5  & 59.8 & 45.4 & 56.5 & 81.1 & 71.0   & 84.0     & 85.5 & 92.6 \\
70     & 67.3 & 52.1 & 71.9 & 82.8 & 81.0    & 60.2 & 45.1 & 56.3 & 81.6 & 70.6 & 83.8     & 85.7 & 92.8 \\
80     & 67.5 & 51.9 & 71.9 & 83.2 & 80.9  & 60.3 & 45.5 & 56.3 & 82.8 & 70.1 & 84.0     & 85.7 & 93.0 \\
90     & 67.6 & 51.8 & 71.6 & 83.1 & 81.3  & 60.7 & 45.4 & 56.1 & 83.7 & 70.2 & 83.9     & 85.4 & 93.8 \\
100    & 67.3 & 51.8 & 71.3 & 83.3 & 81.1  & 60.8 & 45.5 & 55.8 & 82.9 & 70.0 & 84.1     & 85.7 & 94.7 \\ \bottomrule
\end{tabular}}
\label{tabs:iteration}
\end{table*}

\begin{table*}[!ht]
    \centering
    \caption{Attack success rate (\%) across ten models on adversarial examples crafted on ResNet-18 by different operation number}
    \resizebox{\textwidth}{!}{
    \begin{tabular}{cccccccccccc}
    \hline
        Operation Number & Res-18 & Res-101 & NeXT-50 & Denset-121 & Inc-v3 & Inc-v4 & ViT & PiT & Visformer & Swin & Average \\ \hline
        1 & 100.0 & 96.7 & 96.9 & 98.3 & 90.7 & 89.9 & 46.6 & 56.5 & 74.6 & 76.1 & 82.6 \\ 
        2 & 100.0 & 99.3 & 99.2 & 99.6 & 96.9 & 97.4 & 63.7 & 71.1 & 86.6 & 86.0 & 90.0 \\ 
        3 & 100.0 & 99.4 & 99.5 & 99.6 & 98.2 & 98.6 & 63.2 & 76.0 & 89.1 & 89.5 & 91.2 \\
        4 & 100.0 & 99.6 & 99.6 & 99.8 & 98.5 & 99.4 & 64.1 & 77.1 & 90.1 & 90.0 & 91.8 \\ 
        5 & 100.0 & 99.6 & 99.7 & 99.8 & 98.6 & 99.5 & 64.9 & 77.8 & 90.5 & 90.3 & 92.0 \\ \hline
    \end{tabular}}
\label{tabs:ops_num}
\end{table*}
%\subsection{Visualization for different attacking methods}

\begin{table*}[]
\centering
\caption{Attack success rate (\%) across ten models on adversarial examples generated on Res-18 by different number of samples.}
\resizebox{\linewidth}{!}{
\begin{tabular}{cccccccccccc}
\toprule
Sample Number & Res-18 & Res-101 & NeXt-50 & Dense-121 & Inc-v3 & Inc-v4 & ViT  & PiT  & Visformer & Swin & Average \\ \midrule
1             & 100.0  & 90.6    & 92.3    & 95.3      & 85.5   & 82.5   & 38.9 & 46.4 & 61.0      & 64.9 & 75.7    \\
2             & 100.0  & 95.4    & 95.7    & 98.0      & 91.3   & 90.0   & 47.9 & 55.9 & 72.7      & 74.1 & 82.1    \\
3             & 100.0  & 96.7    & 97.1    & 98.6      & 93.1   & 93.4   & 51.6 & 59.4 & 78.6      & 77.7 & 84.6    \\
4             & 100.0  & 97.3    & 98.3    & 98.9      & 94.4   & 94.0   & 55.3 & 62.7 & 79.0      & 80.7 & 86.1    \\
5             & 100.0  & 98.3    & 98.3    & 99.4      & 95.4   & 95.1   & 57.4 & 65.7 & 82.6      & 83.1 & 87.5    \\
6             & 100.0  & 99.1    & 98.7    & 99.6      & 96.0   & 96.5   & 59.3 & 67.2 & 83.1      & 82.2 & 88.2    \\
7             & 100.0  & 99.3    & 98.4    & 99.6      & 96.1   & 96.3   & 61.2 & 67.9 & 85.0      & 83.5 & 88.7    \\
8             & 100.0  & 99.1    & 98.9    & 99.6      & 97.2   & 96.0   & 59.5 & 68.9 & 84.4      & 85.1 & 88.9    \\
9             & 100.0  & 99.2    & 99.2    & 99.5      & 97.0   & 96.4   & 62.3 & 70.5 & 86.3      & 86.3 & 89.7    \\
10            & 100.0  & 99.3    & 99.2    & 99.6      & 96.9   & 97.4   & 63.7 & 71.1 & 86.6      & 86.0 & 90.0    \\
11            & 100.0  & 99.2    & 99.0    & 99.7      & 96.5   & 97.2   & 64.7 & 72.7 & 87.1      & 86.5 & 90.3    \\
12            & 100.0  & 99.1    & 98.8    & 99.8      & 96.7   & 96.6   & 63.8 & 72.7 & 86.6      & 86.0 & 90.0    \\
13            & 100.0  & 99.3    & 99.0    & 99.7      & 96.0   & 97.5   & 65.4 & 72.1 & 87.6      & 86.7 & 90.3    \\
14            & 100.0  & 99.4    & 99.4    & 99.6      & 96.9   & 97.2   & 65.4 & 73.8 & 88.5      & 89.2 & 90.9    \\
15            & 100.0  & 99.2    & 99.5    & 99.6      & 97.3   & 97.5   & 65.4 & 73.0 & 88.1      & 86.8 & 90.6    \\
16            & 100.0  & 99.3    & 99.4    & 99.7      & 97.4   & 97.6   & 67.2 & 74.7 & 88.6      & 87.8 & 91.2    \\
17            & 100.0  & 99.4    & 99.3    & 99.7      & 97.9   & 98.1   & 66.4 & 73.0 & 89.1      & 87.9 & 91.1    \\
18            & 100.0  & 99.2    & 99.3    & 99.5      & 97.2   & 97.3   & 66.7 & 74.5 & 89.3      & 88.1 & 91.0    \\
19            & 100.0  & 99.3    & 99.2    & 99.6      & 97.4   & 97.9   & 66.1 & 73.9 & 88.4      & 87.9 & 91.1    \\
20            & 100.0  & 99.3    & 99.6    & 99.7      & 96.6   & 97.5   & 66.4 & 74.2 & 88.8      & 89.3 & 91.1    \\
21            & 100.0  & 99.4    & 99.4    & 99.5      & 97.0   & 98.2   & 66.1 & 75.0 & 89.0      & 87.8 & 91.1    \\
22            & 100.0  & 99.3    & 99.6    & 99.7      & 97.0   & 97.8   & 67.8 & 75.0 & 89.3      & 88.8 & 91.4    \\
23            & 100.0  & 99.4    & 99.3    & 99.6      & 97.0   & 98.0   & 68.3 & 74.2 & 89.6      & 88.9 & 91.4    \\
24            & 100.0  & 99.5    & 99.4    & 99.7      & 97.6   & 97.9   & 67.4 & 75.4 & 89.6      & 89.7 & 91.6    \\
25            & 100.0  & 99.3    & 99.5    & 99.5      & 97.4   & 98.1   & 67.3 & 75.1 & 88.8      & 88.4 & 91.3    \\
26            & 100.0  & 99.3    & 99.4    & 99.6      & 97.3   & 98.5   & 68.1 & 76.1 & 89.6      & 88.9 & 91.7    \\
27            & 100.0  & 99.4    & 99.4    & 99.8      & 97.6   & 97.7   & 67.7 & 76.3 & 90.0      & 89.7 & 91.8    \\
28            & 100.0  & 99.3    & 99.2    & 99.8      & 97.6   & 98.0   & 68.4 & 76.8 & 90.3      & 89.6 & 91.9    \\
29            & 100.0  & 99.3    & 99.4    & 99.6      & 97.5   & 98.4   & 67.8 & 75.5 & 89.5      & 89.8 & 91.7    \\
30            & 100.0  & 99.4    & 99.6    & 99.6      & 97.6   & 98.4   & 68.3 & 76.1 & 90.3      & 88.7 & 91.8    \\
31            & 100.0  & 99.5    & 99.5    & 99.6      & 97.5   & 98.4   & 68.2 & 76.2 & 89.7      & 90.4 & 91.9    \\
32            & 100.0  & 99.5    & 99.5    & 99.5      & 98.0   & 98.4   & 68.6 & 75.9 & 90.2      & 89.5 & 91.9    \\
33            & 100.0  & 99.3    & 99.5    & 99.7      & 97.6   & 98.4   & 68.0 & 76.6 & 90.2      & 90.1 & 91.9    \\
34            & 100.0  & 99.5    & 99.5    & 99.8      & 97.9   & 98.2   & 69.3 & 76.7 & 90.4      & 90.2 & 92.2    \\
35            & 100.0  & 99.5    & 99.4    & 99.8      & 98.0   & 98.8   & 69.9 & 76.6 & 90.3      & 90.2 & 92.2    \\
36            & 100.0  & 99.4    & 99.6    & 99.8      & 97.7   & 98.2   & 70.1 & 76.9 & 90.0      & 90.1 & 92.2    \\
37            & 100.0  & 99.6    & 99.6    & 99.8      & 97.6   & 98.2   & 68.8 & 76.9 & 90.6      & 90.6 & 92.2    \\
38            & 100.0  & 99.4    & 99.5    & 99.8      & 97.6   & 98.3   & 69.5 & 76.0 & 91.3      & 89.8 & 92.1    \\
39            & 100.0  & 99.4    & 99.4    & 99.5      & 97.3   & 98.1   & 70.5 & 77.8 & 90.6      & 90.2 & 92.3    \\
40            & 100.0  & 99.3    & 99.6    & 99.8      & 97.9   & 98.6   & 67.7 & 76.1 & 90.4      & 90.0 & 91.9    \\
41            & 100.0  & 99.5    & 99.6    & 99.7      & 97.6   & 98.5   & 69.0 & 77.4 & 90.4      & 90.8 & 92.2    \\
42            & 100.0  & 99.5    & 99.6    & 99.8      & 97.6   & 98.4   & 69.7 & 76.5 & 90.7      & 90.2 & 92.2    \\
43            & 100.0  & 99.5    & 99.3    & 99.7      & 98.0   & 98.8   & 70.1 & 77.2 & 91.3      & 89.7 & 92.4    \\
44            & 100.0  & 99.5    & 99.6    & 99.8      & 98.2   & 98.3   & 69.5 & 76.6 & 90.3      & 89.8 & 92.2    \\
45            & 100.0  & 99.6    & 99.6    & 99.8      & 97.7   & 98.4   & 69.7 & 77.2 & 90.6      & 90.4 & 92.3    \\
46            & 100.0  & 99.5    & 99.7    & 99.8      & 97.7   & 98.5   & 69.6 & 77.1 & 91.6      & 90.4 & 92.4    \\
47            & 100.0  & 99.7    & 99.8    & 99.8      & 97.9   & 98.9   & 69.9 & 77.0 & 91.4      & 90.9 & 92.5    \\
48            & 100.0  & 99.5    & 99.5    & 99.7      & 97.6   & 98.4   & 69.5 & 76.9 & 90.9      & 91.3 & 92.3    \\
49            & 100.0  & 99.6    & 99.6    & 99.8      & 97.8   & 98.7   & 69.9 & 76.9 & 91.3      & 90.8 & 92.2    \\
50            & 100.0  & 99.5    & 99.5    & 99.8      & 98.2   & 98.6   & 69.7 & 77.4 & 91.5      & 91.4 & 92.6   \\ \bottomrule
\end{tabular}}
\label{tabs:samples}
\end{table*}

%\subsection{Counting on difference transformation been used}

\end{document}